\documentclass[10pt,a4paper]{article}

\usepackage[T1]{fontenc}
\usepackage[utf8]{inputenc}
\usepackage[margin=0.66in,top=0.71in,bottom=0.72in]{geometry}
\usepackage{amsmath,amssymb,amsthm,mathtools}
\usepackage{graphicx,booktabs,array}
\usepackage{microtype}
\usepackage[hidelinks]{hyperref}
\hypersetup{pdftitle={Tangent incompleteness under coupled image transformations: margin reconstruction and decision consequences},
pdfauthor={Karim Bounja and Abdeljalil Sakat},
pdfkeywords={decision margin, tangent approximation, bilinear resampling, coupled transformations, geometric robustness}}
\usepackage{placeins}
\usepackage{dblfloatfix}

\graphicspath{{figures/}}
\newcommand{\resultlead}[1]{\par\smallskip\noindent\emph{#1.}\enspace}
\newtheorem{theorem}{Theorem}
\newtheorem{proposition}[theorem]{Proposition}
\newtheorem{corollary}[theorem]{Corollary}
\theoremstyle{remark}
\newtheorem*{remark}{Remark}
\newcommand{\E}{\mathbb E}
\newcommand{\Var}{\operatorname{Var}}
\newcommand{\R}{\mathbb R}

\begin{document}
\twocolumn[{%
\begin{minipage}{\textwidth}
\centering
{\LARGE\bfseries Tangent incompleteness under coupled image transformations:\\[3pt]
margin reconstruction and decision consequences\par}
\vspace{9pt}
{\large Karim Bounja$^{*}$ \qquad Abdeljalil Sakat\par}
\vspace{3pt}
{\small Laboratory for Analysis and Modeling of Systems and Decision Support (LAMSAD),\\
Hassan First University of Settat, Morocco\par}
\vspace{3pt}
{\small $^{*}$Corresponding author: \href{mailto:k.bounja.doc@uhp.ac.ma}{k.bounja.doc@uhp.ac.ma}\par}
\vspace{5pt}
{\small\textit{Preprint}\par}
\vspace{12pt}
\begin{minipage}{0.98\textwidth}
\noindent\textbf{Abstract}\par\vspace{2pt}
\small
Can accurate responses to individual geometric image transformations
recover their combined effect on a classifier's decision? For
identity-aligned bilinear resampling, coordinatewise assembly may fail
already at first order. Projecting the implemented transformation onto
a fixed logit-margin gradient, we derive an explicit response comprising
a linear term and signed absolute-value ridges. An exact even--odd
identity and ridge-cancellation conditions characterize the missing
information. Under independent parameter laws, a noncancelling
off-coordinate ridge obstructs even unrestricted additive fits. The coupled
correction has $O(s^2)$ projected error, versus a possible
$\Theta(s)$ axiswise defect; transfer to realized network margins
requires local smoothness. Projected RMS errors fall by $82.96\%$ on
CIFAR-10 and $82.38\%$ on ImageNetV2 relative to axiswise assembly;
realized fixed-pair margin RMS errors fall by $39.75\%$ and $42.73\%$.
The representation prospectively identifies nine strict nonadditive
network-margin sign configurations and supports exact minimization of
a three-parameter threat surrogate, with conditional selection-regret
guarantees. In an independent CIFAR-10 search, coupled-selected
transformations uncover 51 top-1 changes versus 41 for axes. More
directly, on 256 independent ImageNetV2/ConvNeXt-Tiny images and
16,384 transformed decisions at the prespecified scale $s=.40$,
multiclass top-1 flip-prediction errors decrease from $11.97\%$ to
$10.23\%$ (paired improvement $1.74$ percentage points; 95\% interval
$[0.93,2.61]$). Directional finite differences are slightly more
accurate ($10.00\%$) at higher measured model-construction cost.
These results connect a concrete first-order sampling interaction to
margin reconstruction, decision prediction, and threat selection,
without claiming finite-scale network certification.
\vspace{7pt}
\end{minipage}
\end{minipage}
\vspace{14pt}
}]

\section{Introduction}

Classical transformation tangents describe local changes of digital
images under geometric perturbations \cite{simard1991,simard2000}.
For a differentiable transformation map, the first-order joint
response is determined by its coordinate derivatives. An
implemented image transformation also includes resampling, however,
and identity-aligned bilinear interpolation has different one-sided
slopes at pixel-grid nodes. Consequently, assembling individually
corrected rotation and translation responses may miss first-order
information about their simultaneous action.

We study this question after projection onto a \emph{fixed classifier
margin}, rather than in image space alone. For a clean input $x$,
logit margin $m=f_k-f_j$, and gradient $g=\nabla m(x)$, let
$Z_s(U)=g^\top[T_{sU}(x)-x]$ denote the projected response to an
implemented transformation. We establish an exact even--odd error
identity and, for bilinear resampling, the first-order expansion
\begin{equation}
\begin{aligned}
Z_s(U)&=s\{\ell^\top U+H(U)\}+O_{L^2}(s^2),\\
H(U)&=\sum_n C_n|n^\top U|.
\end{aligned}
\label{eq:intro-ridges}
\end{equation}
The signed coefficients depend on image second differences and the
margin gradient, so nonsmooth image responses can cancel after
projection. We characterize this cancellation and identify
noncoordinate ridges that obstruct axiswise assembly and even
unrestricted coordinate-additive $L^2$ fits under independent
parameter laws. Retaining the joint ridges yields $O(s^2)$ projected
error, compared with a possible $\Theta(s)$ axiswise error; transfer
to actual network margins is conditional on local regularity.

We also distinguish reconstruction from \emph{decision consequences}.
Sampler-derived ridges produce fixed-pair sign patterns incompatible
with additive decision models, admit exact minimization of the studied
three-parameter rotation--translation surrogate, and give conditional
fixed-pair margin selection-regret bounds. The accompanying studies
assess projected and realized-margin errors, prospective nonadditive
witnesses, threat selection, and binary multiclass top-1 event
prediction on CIFAR-10, SVHN and ImageNetV2. In an independent
ImageNetV2/ConvNeXt-Tiny confirmation, coupled prediction improves on
axiswise assembly, while directional finite differences remain a
slightly more accurate comparator. We do not claim finite-scale
network certification.

\section{Relation to prior work}

\emph{Transformation models and sampling.}
Tangent distance and propagation motivate the local geometric model
\cite{simard1991,simard2000}. Numerical interpolation affects tangent
linearization \cite{polavarapu1996,tanguay1997}; differentiable
sampling and linearized multi-sampling are also studied in computer
vision \cite{jaderberg2015,jiang2019}. Our object is the
\emph{inference-time bilinear sampler projected onto a trained
classifier's fixed logit-margin gradient}.

\emph{Additivity and nonsmooth structure.}
Coordinate decompositions, functional ANOVA, abs-linear analysis and
ridge-function theory provide the relevant mathematical background
\cite{rabitzalis1999,hoeffding1948,griewank2013,pinkus2013}.
Four-corner and finite-set additive-sign obstructions are classical
\cite{riguet1951,freixas2017}. Here these tools are applied to the
explicit signed ridge coefficients generated by the implemented
sampler, including their margin-dependent cancellations.

\emph{Geometric robustness and decisions.}
Geometric attacks and robustness verification address transformed
network decisions and the effects of interpolation and composition
\cite{engstrom2019,balunovic2019,fischer2020,batten2024,wang2023georobust}.
Our complementary result isolates the \emph{first-order joint
margin interaction} missing from coordinate assembly and evaluates
its consequences for local prediction and threat selection. The
conditional margin-regret bound is not a nonlinear-network
robustness certificate.

\section{Projected tangent error}
\label{sec:projected-tangent}

Let $f:\R^d\to\R^K$ denote classifier logits. Fix an input $x$ with predicted class $k$, a competitor $j\ne k$, and a differentiable positive logit margin $m(z)=f_k(z)-f_j(z)$. Set $g=\nabla m(x)$; both $j$ and $g$ remain fixed. For the \emph{implemented} transformation $T_{sU}$, write
\begin{equation}
\delta_s(U)=T_{sU}(x)-x,\qquad
Z_s(U)=g^\top\delta_s(U).
\label{eq:proj}
\end{equation}
Here $s>0$ scales the transform and $U\in\R^q$ has a centrally symmetric law, $U\overset d=-U$. 
Central symmetry permits exact even--odd orthogonality
without restricting the transformation family: any
direction law can be symmetrized by random sign reversal,
preserving its directional support up to sign. A prescribed linear tangent predicts
$\widehat Z_{a,s}=s a^\top U$.
Define the even and odd parts
$C_s=[Z_s(U)+Z_s(-U)]/2$ and
$S_s=[Z_s(U)-Z_s(-U)]/2$, and let $b_s=\E Z_s$.
The quantity $b_s=\E Z_s=g^\top\E\delta_s(U)$
measures the mean projected margin drift.
This drift can be nonzero even under centrally symmetric
transformation directions, since the realized perturbation
$\delta_s(U)$ is not necessarily odd in $U$.

A parameter-linear tangent is necessarily odd in $U$,
whereas the realized projected response is not constrained to be odd.
Its even component may contribute to systematic drift
and variance, raising the question of first-order
completeness: $\|Z_s-sa^\top U\|_{L^2}=o(s)$.
While linearity of directional derivatives characterizes
Hadamard differentiability under suitable conditions
\cite{shapiro1990,fangsantos2019}, our setting concerns
the $L^2(P_U)$ completeness of a prescribed tangent after
projection onto a fixed decision margin.
Using parity orthogonality, Theorem~\ref{thm:parity}
establishes an exact error decomposition and a
necessary-and-sufficient criterion. The contribution
lies in this margin-specific characterization, which
separates irreducible even error from odd tangent mismatch
and accounts for possible cancellation of image-space
nonsmoothness under projection.

\begin{theorem}[Exact error and first-order completeness]
\label{thm:parity}
If $Z_s\in L^2$ and $U\in L^2(\R^q)$, then for every $a\in\R^q$,
\begin{equation}
\boxed{
\E|Z_s-sa^\top U|^2
=b_s^2+\Var(C_s)
+\E|S_s-sa^\top U|^2.
}
\label{eq:exact}
\end{equation}
Suppose additionally that $\delta_s/s\to D_x(U)$ in $L^2$ as $s\downarrow0$. Writing $h(U)=g^\top D_x(U)$ and $h_e,h_o$ for its parity components,
\begin{equation}
\lim_{s\downarrow0}
\frac{\E|Z_s-sa^\top U|^2}{s^2}
=\E|h_e|^2+\E|h_o-a^\top U|^2.
\label{eq:firstorder}
\end{equation}
Thus $\|Z_s-sa^\top U\|_{L^2}=o(s)$ if and only if $h_e=0$ and $h_o=a^\top U$ almost surely. In particular, every parameter-linear surrogate has error at least $[b_s^2+\Var(C_s)]^{1/2}$ in projected root mean square.
\end{theorem}

\begin{proof}
Since $U\overset d=-U$, even and odd functions of $U$
are orthogonal in $L^2$. In particular,
$\E C_s=\E Z_s=b_s$. As $a^\top U$ is odd,
Pythagoras gives
\[
\E|Z_s-sa^\top U|^2
=\E C_s^2+\E|S_s-sa^\top U|^2,
\]
and $\E C_s^2=b_s^2+\Var(C_s)$, proving
\eqref{eq:exact}.

Next, since $g$ is fixed, Cauchy--Schwarz yields
\[
\left\|\frac{Z_s}{s}-h\right\|_{L^2}
\leq \|g\|
\left\|\frac{\delta_s}{s}-D_x\right\|_{L^2}
\longrightarrow 0.
\]
By continuity of the $L^2$ norm and orthogonality
of the even and odd parts of $h$,
\[
\begin{aligned}
\lim_{s\downarrow0}\frac{\E|Z_s-sa^\top U|^2}{s^2}
&=\|h-a^\top U\|_{L^2}^2\\
&=\E|h_e|^2+\E|h_o-a^\top U|^2,
\end{aligned}
\]
which proves \eqref{eq:firstorder}.
Since both terms are nonnegative, the limit
vanishes exactly when $h_e=0$ and
$h_o=a^\top U$ almost surely.
The stated lower bound follows directly
from \eqref{eq:exact}.
\end{proof}

Theorem~\ref{thm:parity} characterizes the first-order
$L^2$ completeness of a prescribed tangent:
$h_e=0$ and $h_o=a^\top U$ almost surely.
The criterion is margin-specific: since
$h_e=g^\top D_{x,e}$ and $\ker(g^\top)\neq\{0\}$
for $d>1$, a nonzero even image-space response
may vanish after projection.

The even-part operator is contractive in $L^2$,
so $Z_s/s\to h$ implies $C_s/s\to h_e$.
Since $sa^\top U$ and $S_s$ are odd, symmetry of $U$
makes their difference orthogonal to the even part
$C_s$ in $L^2$. Thus \eqref{eq:exact} yields
\[
\begin{aligned}
\inf_{a\in\R^q}\|Z_s-sa^\top U\|_{L^2}
&\geq \|C_s\|_{L^2}\\
&=\bigl(b_s^2+\Var(C_s)\bigr)^{1/2}\\
&=s\|h_e\|_{L^2}+o(s).
\end{aligned}
\]
Thus $h_e\neq0$ precludes first-order completeness. The exact decomposition \eqref{eq:exact} separates
this irreducible even defect from odd tangent mismatch.
In particular, equality of projected variances
alone cannot establish the $L^2$ completeness
characterized by \eqref{eq:firstorder}. A nonzero first-order even response therefore reveals
a structural limitation of parameter-linear tangents:
being odd in $U$, they cannot represent this component
or achieve $o(s)$ projected RMS error, regardless
of how their coefficients are fitted.

\section{Explicit bilinear correction}

We now specialize the criterion to bilinear resampling,
where changes in interpolation slope at grid nodes
can produce first-order linearization errors
\cite{polavarapu1996,tanguay1997}.
By Theorem~\ref{thm:parity}, an even first-order
component surviving margin projection obstructs
parameter-linear tangent completeness.
Under joint transformations, as studied in geometric
robustness \cite{balunovic2019,batten2024}, such terms
may escape reconstruction from separate responses.
We characterize this coupled discrepancy and
construct a correction.

Consider bilinear image resampling \cite{wolberg1990,jaderberg2015}
on a finite identity-aligned pixel grid, with fixed padding
and clipping absent or locally inactive. Let $\widetilde x$ denote the
channelwise extension prescribed by the padding rule,
and $B_{\widetilde x}$ its bilinear interpolant.
At pixel $p$, define
$\nabla_i^c\widetilde x_p=
(\widetilde x_{p+e_i}-\widetilde x_{p-e_i})/2$
and
$\Delta_{ii}\widetilde x_p=
\widetilde x_{p+e_i}-2\widetilde x_p+
\widetilde x_{p-e_i}$,
for $i\in\{x,y\}$.
Displacements $h$ are measured in pixel-grid
coordinates, after conversion from normalized
sampler coordinates when applicable.
At an integer grid node, for $|h_x|,|h_y|<1$,
\begin{align}
B_{\widetilde x}(p+h)-\widetilde x_p
={}&h_x\nabla_x^c\widetilde x_p
+\tfrac12|h_x|\Delta_{xx}\widetilde x_p
\nonumber\\
&+h_y\nabla_y^c\widetilde x_p
+\tfrac12|h_y|\Delta_{yy}\widetilde x_p\\
&+O(|h_xh_y|).
\label{eq:bilinear}
\end{align}
Expanding the bilinear weights in each adjacent
cell leaves a mixed finite-difference term bounded
by a constant depending on $\widetilde x$ times
$|h_xh_y|$. Along either sampling-coordinate axis
this term vanishes, giving the exact identity,
for $|t|<1$,
\[
B_{\widetilde x}(p+te_i)-\widetilde x_p
=t\nabla_i^c\widetilde x_p
+\tfrac12|t|\Delta_{ii}\widetilde x_p.
\]

The absolute-value terms in \eqref{eq:bilinear}
contribute at first order, whereas the mixed
remainder is $O(\|h\|^2)$.
For a joint horizontal displacement
$h_x=s\sum_{r=1}^q v_rU_r+O(s^2)$,
the joint and axiswise-assembled contributions
differ, to leading order, by
\[
\frac{s}{2}\Delta_{xx}\widetilde x_p
\left(
\left|\sum_{r=1}^q v_rU_r\right|
-\sum_{r=1}^q|v_rU_r|
\right).
\]
This leading term is nonzero when
$\Delta_{xx}\widetilde x_p\neq0$ and the signed
displacement contributions include both positive
and negative values. Whether the discrepancy
survives margin projection depends on cancellations
across pixels, examined below.

\subsection{Scalar transforms}

Suppose the sampling-coordinate displacement satisfies
$h_p(t)=tv_p+O(t^2)$ uniformly in $p$ as $t\to0$.
Define
\begin{align}
L={}&\sum_p g_p^\top
\bigl(v_{p,x}\nabla_x^c\widetilde x_p
+v_{p,y}\nabla_y^c\widetilde x_p\bigr),
\nonumber\\
K={}&\frac12\sum_p g_p^\top
\bigl(|v_{p,x}|\Delta_{xx}\widetilde x_p
+|v_{p,y}|\Delta_{yy}\widetilde x_p\bigr).
\label{eq:LK}
\end{align}
For bounded scalar $U$, we have
$h_{p,i}(sU)=sUv_{p,i}+O(s^2)$ and
$|h_{p,i}(sU)|=s|U||v_{p,i}|+O(s^2)$.
Substitution into \eqref{eq:bilinear}, followed
by projection onto $g$, therefore gives
\begin{equation}
Z_s(U)=s\{LU+K|U|\}+O_{L^2}(s^2).
\label{eq:scalar}
\end{equation}
Writing \eqref{eq:scalar} as
$Z_s(U)=s(LU+K|U|)+R_s(U)$, where
$\|R_s\|_{L^2}=O(s^2)$, the corrected surrogate
$\widehat Z_{\mathrm{corr},s}=s(LU+K|U|)$
satisfies
\[
\|Z_s-\widehat Z_{\mathrm{corr},s}\|_{L^2}
=\|R_s\|_{L^2}=O(s^2).
\]

For $U\sim\mathrm{Unif}[-1,1]$, symmetry yields
$\E[U|U|]=0$ and $\E[U^2]=1/3$. Hence
\[
\E|s(LU+K|U|)-saU|^2
=\frac{s^2}{3}\bigl((L-a)^2+K^2\bigr).
\]
The leading term is minimized at $a=L$,
and the $O_{L^2}(s^2)$ remainder does not
affect the leading-order minimum. Thus
\begin{equation}
\inf_a\E|Z_s-saU|^2
=\tfrac{K^2}{3}s^2+o(s^2).
\label{eq:scalarbest}
\end{equation}

The coefficient $K$ may vanish through
cancellation after margin projection, even
when individual pixels have nonzero second
differences. For the selected margin, $K\neq0$
therefore prevents first-order completeness
of any parameter-linear tangent, whereas the
explicit even correction achieves $O(s^2)$
projected RMS error.

Having corrected the first-order even response
missed by parameter-linear tangents in the
scalar case, a natural strategy for simultaneous
transformations is to assemble the individually
corrected responses. This is justified at first
order for differentiable parameter-to-image
maps, as in classical tangent-based approaches
\cite{simard1991,simard2000}. However, bilinear resampling at grid nodes
is not necessarily differentiable with respect
to transformation parameters, as reflected
by the absolute-value terms in \eqref{eq:bilinear}.

Related work has shown that interpolation and
rounding can prevent exact composition of
digital image transformations \cite{fischer2020},
while geometric robustness verification handles
composed transformations through sound relaxations
\cite{balunovic2019,batten2024}.
Our focus is different: determining whether
individually accurate responses suffice to
reconstruct the joint response projected onto
a fixed logit-margin gradient, quantifying
the leading assembly error and its cancellation,
and constructing an explicit correction using
the clean image, its margin gradient, and
the implemented sampler.

\begin{figure*}[!t]
\centering
\includegraphics[width=.84\textwidth]{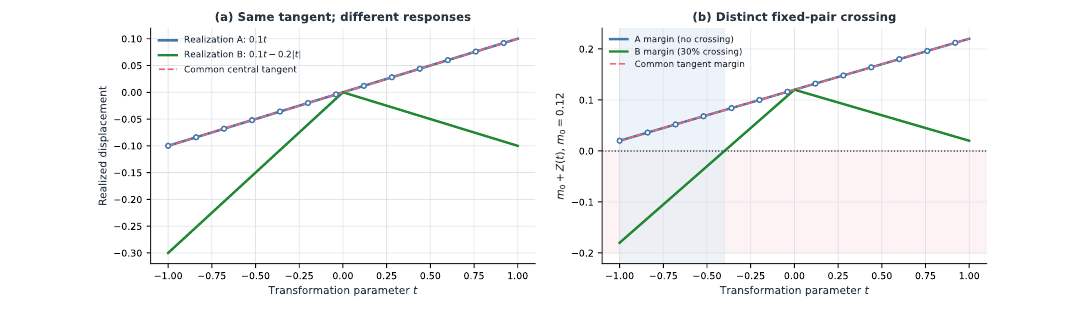}
\caption{Exact three-pixel example: two interpolated responses share the tangent $0.1t$, but one contains $-0.2|t|$. At clean margin $0.12$ and $t\sim\mathrm{Unif}[-1,1]$, their projected crossing probabilities are $0$ and $0.30$. No trained classifier is used.}
\label{fig:toy}
\end{figure*}

\subsection{Simultaneous transformations}

Let $U\in\R^q$ have bounded support, and assume that the sampling displacement at each pixel $p$ satisfies
$h_p(sU)=sV_pU+O(s^2)$ uniformly in $p$ and $U$, with $V_p\in\R^{2\times q}$.
Writing $V_{p,x}$ and $V_{p,y}$ for its rows, define
\begin{align}
\ell^\top U={}&\sum_p g_p^\top\bigl[
(V_{p,x}U)\nabla_x^c\widetilde x_p
+(V_{p,y}U)\nabla_y^c\widetilde x_p\bigr],
\nonumber\\
H(U)={}&\frac12\sum_p g_p^\top\bigl[
|V_{p,x}U|\Delta_{xx}\widetilde x_p
+|V_{p,y}U|\Delta_{yy}\widetilde x_p\bigr].
\label{eq:joint}
\end{align}

The linear term $\ell^\top U$ is odd and
reconstructible from its coordinate responses.
In contrast, $H$ is even, positively homogeneous,
and a signed sum of ridge terms $|V_{p,i}U|$,
which may couple several transformation
parameters. Its coordinate responses may
therefore fail to determine its joint response:
for $H(U)=|U_1+U_2|$, axiswise assembly gives
$|U_1|+|U_2|$, agreeing on both axes but
differing at $U=(1,-1)$. Since $H$ enters
$Z_s$ at order $s$, a nonzero assembly
discrepancy in $L^2(P_U)$ produces a
first-order error.
The following proposition quantifies this defect
and establishes the accuracy of a coupled correction.


\begin{proposition}[Coupled response versus axiswise assembly]
\label{prop:coupled}
Under the above assumptions,
\begin{equation}
Z_s(U)=s\{\ell^\top U+H(U)\}+r_s(U),
\quad \|r_s\|_{L^2}=O(s^2).
\label{eq:jointresult}
\end{equation}
For the coordinate vectors $e_r$, set $K_r=H(e_r)$ and define the axiswise-assembled surrogate
$\widehat Z_{\mathrm{ax},s}
=s(\ell^\top U+\sum_rK_r|U_r|)$.
Then
\begin{equation}
\begin{split}
\bigl|\|Z_s-\widehat Z_{\mathrm{ax},s}\|_{L^2}
-sd_{\mathrm{ax}}\bigr|&=O(s^2),\\
d_{\mathrm{ax}}&=
\Bigl\|H-\sum_rK_r|U_r|\Bigr\|_{L^2}.
\end{split}
\label{eq:axerr}
\end{equation}
The coupled surrogate
$\widehat Z_{\mathrm{joint},s}=s(\ell^\top U+H(U))$
achieves $O(s^2)$ projected RMS error.
If $U\overset d=-U$, then
\begin{equation}
\inf_{a\in\R^q}\E|Z_s-sa^\top U|^2
=s^2\E H(U)^2+O(s^3).
\label{eq:jointoptimal}
\end{equation}
\end{proposition}

\begin{proof}
Substitute $h_p(sU)=sV_pU+O(s^2)$ into \eqref{eq:bilinear} and project onto $g$. The absolute-value map is Lipschitz, and bounded $U$ and the finite grid give the $L^2$ remainder in \eqref{eq:jointresult}. Subtracting the axiswise surrogate yields
\[
Z_s-\widehat Z_{\mathrm{ax},s}
=s\Bigl(H(U)-\sum_rK_r|U_r|\Bigr)+r_s,
\]
so the reverse triangle inequality proves \eqref{eq:axerr}.
For \eqref{eq:jointoptimal}, let
$\mathcal L=\{a^\top U:a\in\R^q\}$.
Under symmetry, $H$ is orthogonal to $\mathcal L$, while $\ell^\top U\in\mathcal L$.
Since distance to a set is $1$-Lipschitz,
\[
\operatorname{dist}_{L^2}(Z_s/s,\mathcal L)
=\|H\|_{L^2}+O(s).
\]
Squaring and multiplying by $s^2$ gives the result.
\end{proof}

Thus $d_{\mathrm{ax}}=0$ exactly characterizes first-order completeness of the axiswise assembly; if $d_{\mathrm{ax}}>0$, its projected RMS error is $\Theta(s)$, versus $O(s^2)$ for the coupled correction. 
Under symmetry, no parameter-linear tangent
can represent a nonzero even component $H$;
even the optimal fit therefore incurs a
$\Theta(s)$ projected RMS error.

\begin{corollary}[Cancellation test]
\label{cor:group}
Assume the law of $U$ charges every nonempty open subset of a neighborhood of $0$. In the finite ridge representation
$H(U)=\sum_i c_i|v_i^\top U|$,
group nonzero vectors $v_i$ that are proportional up to sign. For each unoriented unit direction $n$, set
$C_n=\sum_{i:v_i\parallel n}c_i\|v_i\|$.
Then
\begin{equation}
H(U)=0\ \text{a.s.}\quad\Longleftrightarrow\quad
C_n=0\ \text{for all groups }n.
\label{eq:cancel}
\end{equation}
When the linear tangent matches $\ell^\top U$, this is exactly the criterion for first-order projected completeness.
\end{corollary}

\begin{proof}
Grouping the nonzero ridge vectors in \eqref{eq:joint}
by direction $n=\pm v_i/\|v_i\|$ and using
$|v_i^\top u|=\|v_i\||n^\top u|$ yields
\[
H(u)=\sum_n C_n|n^\top u|,
\qquad
C_n=\sum_{i:v_i\parallel n}c_i\|v_i\|.
\]
By continuity and the support assumption, almost-sure vanishing implies vanishing near $0$. For $q\ge2$, across a generic point of the hyperplane $n^\top u=0$, the gradient jump is $2C_n n$, while the other grouped terms are smooth there; hence $C_n=0$. For $q=1$ there is only one unoriented direction. The converse is immediate.
\end{proof}

Interpolation-induced nonsmoothness may persist at
individual pixels yet cancel after projection onto a
selected margin through the signed gradient weights.
This criterion concerns the entire even response,
not merely its assembly defect: $H=0$ implies
$d_{\mathrm{ax}}=0$, but not conversely. One may therefore try to repair a nonzero $d_{\mathrm{ax}}$ by refitting the coordinate coefficients. The next result identifies the limit of this strategy within a specified finite-dimensional class.

\begin{proposition}[Best affine--absolute coordinate surrogate]
\label{prop:additive}
Assume \eqref{eq:jointresult} and $U\overset d=-U$.
Define
$\mathcal B=\operatorname{span}\{1,|U_1|,\ldots,|U_q|\}\subset L^2(P_U)$
and
$\mathcal A=\{a^\top U+b(U):a\in\R^q,\ b\in\mathcal B\}$.
With $d_{\mathrm{add}}=\operatorname{dist}_{L^2}(H,\mathcal B)$,
\begin{equation}
\left|
\inf_{h\in\mathcal A}\|Z_s-sh\|_{L^2}
-sd_{\mathrm{add}}
\right|=O(s^2).
\label{eq:bestadd}
\end{equation}
If, in addition, $P_U$ charges every nonempty open subset of a neighborhood of $0$ and the ridge coefficients are grouped as in Corollary~\ref{cor:group}, then
\begin{equation}
d_{\mathrm{add}}>0
\quad\Longleftrightarrow\quad
\exists n\not\parallel e_r\ (\forall r): C_n\ne0.
\label{eq:noncoordinate}
\end{equation}
Thus a noncancelling off-coordinate ridge precludes $o(s)$ projected RMS error throughout $\mathcal A$, whereas the coupled correction attains $O(s^2)$.
\end{proposition}

\begin{proof}
Under symmetry, $H$ and $\mathcal B$ are even and orthogonal to all linear forms in $U$. Since $\ell^\top U\in\mathcal A$,
$\operatorname{dist}_{L^2}(\ell^\top U+H,\mathcal A)=d_{\mathrm{add}}$.
Applying the $1$-Lipschitz property of distance to
$Z_s/s=\ell^\top U+H+O_{L^2}(s)$
gives \eqref{eq:bestadd}.
If $d_{\mathrm{add}}=0$, finite dimensionality of $\mathcal B$ gives $H\in\mathcal B$. By continuity and the support condition, this identity holds near $0$; evaluating it at $0$ eliminates the constant term. Across a generic off-coordinate ridge hyperplane, the gradient jump of $H$ is $2C_n n$, while every function in $\mathcal B$ is smooth there. Hence every off-coordinate $C_n$ vanishes. Conversely, if all such coefficients vanish, $H\in\mathcal B$, so $d_{\mathrm{add}}=0$.
\end{proof}

Since $\sum_rK_r|U_r|\in\mathcal B$, we have
$0\le d_{\mathrm{add}}\le d_{\mathrm{ax}}$.
For independent $U_1,U_2\sim\mathrm{Unif}[-1,1]$ and $H(U)=|U_1+U_2|$, the best $L^2$ approximation in $\mathcal B$ is
$\frac16+\frac12(|U_1|+|U_2|)$,
with squared error $13/72>0$.
Thus the first-order obstruction cannot generally be removed by
coefficient fitting in this finite-dimensional class. The next
result rules out \emph{all} square-integrable coordinate-additive
fits under a stronger, explicitly stated parameter-law assumption.

\begin{corollary}[Unrestricted additive obstruction under product laws]
\label{cor:all-additive}
Suppose \eqref{eq:jointresult} holds with $q\ge2$ and $U$ has
independent coordinates whose marginal laws each charge every
nonempty open subinterval of a common $(-\rho,\rho)$.
Let
\[
\mathcal G=\bigl\{c+\sum_{i=1}^q\phi_i(U_i):
\phi_i\in L^2(P_{U_i})\bigr\}.
\]
Set $d_{\rm all}=\operatorname{dist}_{L^2}(H,\mathcal G)$.
The subspace $\mathcal G$ is closed, and its orthogonal
projection satisfies
\begin{equation}
\begin{split}
P_{\mathcal G}H
&=\E H+\sum_{i=1}^q(\E[H\mid U_i]-\E H),\\
d_{\rm all}^2
&=\Var(H)-\sum_{i=1}^q\Var(\E[H\mid U_i]).
\end{split}
\label{eq:all-add-projection}
\end{equation}
With the grouped coefficients $C_n$ of
Corollary~\ref{cor:group},
\begin{equation}
 d_{\rm all}>0\quad\Longleftrightarrow\quad
 \exists\,n\not\parallel e_i\ (\forall i):C_n\ne0.
\label{eq:all-add-obstruction}
\end{equation}
Furthermore,
\[
\inf_{h\in\mathcal G}\|Z_s-sh\|_{L^2}
=s\,d_{\rm all}+O(s^2),\qquad s\downarrow0.
\]
\end{corollary}

\begin{proof}
Independence makes the centered univariate $L^2$ subspaces
mutually orthogonal; their finite sum with the constants is
closed. The orthogonal-projection formula and Pythagoras give
\eqref{eq:all-add-projection}, as in functional ANOVA
\cite{hoeffding1948}. Suppose $d_{\rm all}=0$. Then $H$
agrees almost surely with an additive function. All mixed
coordinate rectangle differences of $H$ therefore vanish
almost surely under the corresponding product law. Continuity
of these differences and full marginal support imply that they
vanish throughout a neighborhood of the origin. Consequently,
for each $i\ne j$, the distribution $\partial_i\partial_jH$
vanishes there. But, since the distinct unoriented ridge
directions have already been grouped,
\[
\partial_i\partial_jH
=2\sum_n C_n n_i n_j\,\delta(n^\top u).
\]
If an off-coordinate group has $C_n\ne0$, choose
$i\ne j$ with $n_i n_j\ne0$ and a generic point of
$n^\top u=0$ on no other grouped ridge hyperplane.
Its nonzero singular measure cannot be canceled, a
contradiction. Conversely, the absence of off-coordinate
groups makes $H$ coordinate-additive, hence $d_{\rm all}=0$.
Finally $\ell^\top U\in\mathcal G$ and distance to a
closed subspace is $1$-Lipschitz; applying
\eqref{eq:jointresult} proves the error expansion.
\end{proof}

For $U_1,U_2$ independent uniform on $[-1,1]$ and
$H(U)=|U_1+U_2|$, the unrestricted optimal fit is
$P_{\mathcal G}H=\frac13+\frac12(U_1^2+U_2^2)$,
with $d_{\rm all}^2=8/45>0$; the finite-dimensional
class above has $d_{\rm add}^2=13/72$.
The unrestricted result is not asserted for dependent
transformation laws. It concerns $L^2$ reconstruction and
does not itself imply a change of classifier decision.

\begin{corollary}[Transfer to the realized fixed-pair margin]
\label{cor:realmargin}
Assume Proposition~\ref{prop:coupled} and
$\|\delta_s(U)\|\le Cs$ uniformly over the support of $U$ for sufficiently small $s$.
Set $\Delta_s(U)=m(T_{sU}(x))-m(x)$.
If $m$ is Fr\'echet differentiable at $x$, then
\begin{equation}
\|\Delta_s-\widehat Z_{\mathrm{joint},s}\|_{L^2}=o(s).
\label{eq:realmargin}
\end{equation}
If $\nabla m$ is locally Lipschitz near $x$, this error is $O(s^2)$. Moreover,
\[
\|\Delta_s-\widehat Z_{\mathrm{ax},s}\|_{L^2}
=s\,d_{\mathrm{ax}}+o(s).
\]
Under the additional symmetry $U\overset d=-U$,
\begin{align}
\inf_{a\in\R^q}\|\Delta_s-sa^\top U\|_{L^2}
&=s\|H\|_{L^2}+o(s),
\label{eq:actual_linear}\\
\inf_{h\in\mathcal A}\|\Delta_s-sh\|_{L^2}
&=s\,d_{\mathrm{add}}+o(s),
\label{eq:actual_additive}
\end{align}
where $\mathcal A$ and $d_{\mathrm{add}}$ are as in Proposition~\ref{prop:additive}.
Under the independent-law assumptions of
Corollary~\ref{cor:all-additive}, also
\begin{equation}
\inf_{h\in\mathcal G}\|\Delta_s-sh\|_{L^2}
=s\,d_{\rm all}+o(s).
\label{eq:real-all-additive}
\end{equation}
Under the locally Lipschitz-gradient assumption, all stated
$o(s)$ remainders improve to $O(s^2)$.
\end{corollary}

\begin{proof}
Fr\'echet differentiability at $x$ gives
$m(x+v)-m(x)=g^\top v+o(\|v\|)$.
The uniform displacement bound implies
$\|\Delta_s-Z_s\|_{L^2}=o(s)$.
If $\nabla m$ is locally Lipschitz with constant $L_m$, then
\[
|\Delta_s-Z_s|\le\frac{L_m}{2}\|\delta_s\|^2,
\]
so the error is $O(s^2)$. Combining these estimates with Proposition~\ref{prop:coupled} gives \eqref{eq:realmargin}; the reverse triangle inequality transfers \eqref{eq:axerr} to $\Delta_s$. The optimal-error statements follow from the $1$-Lipschitz property of distance to the corresponding surrogate classes in $L^2$.
\end{proof}

This transfer applies locally to the same fixed class pair. It does not certify multiclass robustness or finite-scale crossing probabilities; for a piecewise-affine logit model, the locally Lipschitz-gradient case applies within an affine activation region.

\begin{proposition}[Interaction indicator and parity-selective correction]
\label{prop:interaction}
Assume Proposition~\ref{prop:coupled}, evaluating all $L^2$ norms under the same law of $U$. Define
\begin{align}
I_s&=\widehat Z_{\mathrm{joint},s}-\widehat Z_{\mathrm{ax},s},
& D_s(x)&=\|I_s\|_{L^2},\nonumber\\
R_s&=\Delta_s-\widehat Z_{\mathrm{joint},s},
& E_s&=\Delta_s-\widehat Z_{\mathrm{ax},s}.
\label{eq:diagnosticdef}
\end{align}
Then $E_s=I_s+R_s$, $I_s(-U)=I_s(U)$, and
\begin{equation}
D_s(x)=s\,d_{\mathrm{ax}},\qquad
\big|\|E_s\|_{L^2}-D_s(x)\big|
\le\|R_s\|_{L^2}.
\label{eq:diagnosticbound}
\end{equation}
If $G_s=\|E_s\|_{L^2}-\|R_s\|_{L^2}$ denotes the RMS improvement from replacing axiswise assembly by the coupled surrogate, then
\begin{equation}
D_s-2\|R_s\|_{L^2}\le G_s\le D_s.
\label{eq:diagnosticgain}
\end{equation}
When $U\overset d=-U$, the parity components satisfy
\begin{align}
(E_s)_{\mathrm{odd}}&=(R_s)_{\mathrm{odd}},\nonumber\\
\|E_s\|_{L^2}^2-\|R_s\|_{L^2}^2
&=D_s^2+2\E[I_s(R_s)_{\mathrm{even}}].
\label{eq:diagnosticparity}
\end{align}
Replacing $\Delta_s$ by $Z_s$ in these definitions gives $\|R_s\|_{L^2}=O(s^2)$; for $\Delta_s$ the residual is $o(s)$ under Corollary~\ref{cor:realmargin}, and $O(s^2)$ under its locally Lipschitz-gradient assumption.
\end{proposition}

\begin{proof}
Since the surrogates share the odd linear term,
$I_s=s[H(U)-\sum_rK_r|U_r|]$ is even, and $D_s=sd_{\mathrm{ax}}$. The triangle and reverse triangle inequalities applied to $E_s=I_s+R_s$ prove \eqref{eq:diagnosticbound}--\eqref{eq:diagnosticgain}. Under symmetry, even--odd orthogonality gives \eqref{eq:diagnosticparity}. The remainder orders follow from Proposition~\ref{prop:coupled} and Corollary~\ref{cor:realmargin}.
\end{proof}

The indicator $D_s$ depends only on the clean image, margin gradient, sampler geometry, and law of $U$, without transformed-network margin evaluations. Equation~\eqref{eq:diagnosticbound} controls its relation to the realized assembly error, but is not a calibrated confidence bound. At finite $s$, a substantial classifier remainder can obscure the realized error and even the sign of the gain $G_s$. Estimating the diagnostic on directions distinct from those used for evaluation provides a more demanding empirical test.

\begin{remark}[Photometric counterpart]
The nonseparable first-order response identified
above is not specific to spatial resampling.
Without clipping, affine brightness--contrast
transformations $T_{b,c}(x)=(1+c)x+b\mathbf 1$
have an exactly linear projected response,
\[
g^\top(T_{b,c}(x)-x)
=b\,g^\top\mathbf 1+c\,g^\top x,
\]
so axiswise reconstruction is exact.
With clipping, however, a saturated pixel
$x_p=1$ satisfies, for bounded $U=(U_1,U_2)$
and sufficiently small $s>0$,
\[
T_{sU_1,sU_2}(x)_p-x_p
=\frac{s}{2}
\bigl(U_1+U_2-|U_1+U_2|\bigr).
\]
The nonseparable even term can generate an
order-$s$ assembly defect unless it cancels
after margin projection. The parity criterion also applies to photometric
and joint spatial--photometric transformations.
Although the coupled correction is derived here
for bilinear resampling, clipping can generate
analogous first-order interactions, suggesting
extensions to other nonsmooth image operators.
\end{remark}

\resultlead{Interpretation}
The quantities $\|H\|_{L^2}$, $d_{\mathrm{ax}}$,
and $d_{\mathrm{add}}$ measure the leading defects for the best
parameter-linear tangent (under symmetry), prescribed axiswise
assembly, and the finite-dimensional affine--absolute coordinate
surrogates. Under an independent parameter law, $d_{\rm all}$
extends the obstruction to every square-integrable additive fit.
The diagnostic $D_s=s\,d_{\mathrm{ax}}$ isolates omitted coupling;
its transfer to realized margins still depends on the classifier
remainder.

\subsection{Decision consequences of the coupled ridges}
\label{sec:decision-consequences}

A margin-reconstruction defect changes a decision only when it
changes a sign. We distinguish fixed-pair nonadditive witnesses,
realized-margin threat selection, and multiclass top-1 prediction.
The first uses a classical four-corner obstruction; the strongest
local asymptotic guarantee below concerns fixed-pair selection
regret, not top-1 prediction error.

\resultlead{Nonadditive fixed-pair signs}

\begin{proposition}[Nonadditive decision witness]
\label{prop:decision-witness}
Fix all but two transformation coordinates $(r,t)$, and let
$\widehat M_s(r,t)=m(x)+sF(r,t)$ denote the coupled
first-order margin. Suppose that four parameter combinations
$(r_i,t_j)$ satisfy the strict sign pattern
\begin{equation}
\begin{pmatrix}
\widehat M_s(r_1,t_1)&\widehat M_s(r_1,t_2)\\
\widehat M_s(r_2,t_1)&\widehat M_s(r_2,t_2)
\end{pmatrix}
\quad\text{with signs}\quad
\begin{pmatrix}-&+\\+&-\end{pmatrix},
\label{eq:checkerboard}
\end{equation}
or its sign-reversed counterpart. No margin of the form
$c+\phi(r)+\psi(t)$ can reproduce all four signs,
regardless of the univariate functions $\phi,\psi$.
If $\Delta_s$ is the realized margin change and
\begin{equation}
\max_{i,j}|\Delta_s(r_i,t_j)-sF(r_i,t_j)|
<\min_{i,j}|\widehat M_s(r_i,t_j)|,
\label{eq:sign-transfer}
\end{equation}
then the same obstruction holds for the realized fixed-pair
network decisions at those four transformations.
\end{proposition}

\begin{proof}
Every additive margin $A$ satisfies the four-corner identity
$A(r_1,t_1)+A(r_2,t_2)
=A(r_1,t_2)+A(r_2,t_1)$, whereas strictly
negative diagonal entries and strictly positive off-diagonal
entries contradict this equality. The reversed pattern is
identical after changing signs. Condition~\eqref{eq:sign-transfer}
preserves all four signs of the realized margins.
\end{proof}

The four-corner obstruction is classical \cite{riguet1951}.
Here the sampler-derived ridges permit selection of the four
transformations from the clean image and margin gradient,
\emph{before} transformed-network inference. Condition~
\eqref{eq:sign-transfer} certifies sign transfer only if its
remainder is verified. This is a \emph{fixed-pair} pattern, not
a top-1 checkerboard: a negative margin against a competitor forces
a change from the clean winner, but a positive margin does not
rule out another competitor.

For arbitrary finite configurations and any number of
transformation parameters, the corresponding additive obstruction
has the following precise separation characterization. The result
is classical; our application is to sampler-derived margins.

\begin{proposition}[Finite-set additive decision criterion]
\label{prop:finite-additive-sign}
Let $\mathcal S\subset\R^q$ be finite, and let a scalar
function $M:\mathcal S\to\R$ be nonzero at every point, with
both nonempty sets $\mathcal S_+=\{u:M(u)>0\}$ and
$\mathcal S_-=\{u:M(u)<0\}$.
No coordinate-additive function
$A(u)=c+\sum_{i=1}^q\phi_i(u_i)$ reproduces all these signs
\emph{strictly} if and only if there are probability distributions
$P_+$ supported on $\mathcal S_+$ and $P_-$ supported on
$\mathcal S_-$ with the same one-coordinate marginals for
every $i$. For any such pair, let
\begin{equation}
\Lambda(M;P_+,P_-)=\frac12\bigl(
\E_{P_+}M-\E_{P_-}M\bigr)>0.
\label{eq:balanced-lambda}
\end{equation}
Every coordinate-additive $A$ obeys
\begin{equation}
\max_{u\in\mathcal S}|M(u)-A(u)|
\ge\Lambda(M;P_+,P_-).
\label{eq:balanced-margin-witness}
\end{equation}
In particular, if $M=\widehat M_s$ is the coupled fixed-pair
surrogate and the realized network margin differs from it at
every $u\in\mathcal S$ by strictly less than
$\min_{u\in\mathcal S}|\widehat M_s(u)|$, the same strict
sign nonrepresentability holds for the realized fixed-pair
network margins.
\end{proposition}

\begin{proof}
For each coordinate $i$, enumerate the finitely many distinct
values attained on $\mathcal S$. Encode each $u$ by
concatenating the one-hot vectors of its coordinate values.
On this finite set, coordinate-additive functions coincide
with affine functionals of these encodings. Strict sign
representation therefore amounts to strict linear separation
of the convex hulls of the positive and negative encodings.
These compact hulls are strictly separable exactly when they
do not intersect. An intersection amounts to two convex
combinations, one of positive configurations and one of negative
configurations, with identical concatenated coordinate-frequency
vectors; equivalently, $P_+$ and $P_-$ have identical
one-coordinate marginals. For such distributions every additive
$A$ has $\E_{P_+}A=\E_{P_-}A$. Thus
\[
2\Lambda=\E_{P_+}(M-A)-\E_{P_-}(M-A)
\le2\max_{\mathcal S}|M-A|,
\]
which gives \eqref{eq:balanced-margin-witness}. The final
assertion follows because a perturbation strictly smaller
than the minimum absolute surrogate margin preserves all signs.
\end{proof}

The criterion is a classical finite-dimensional separation principle
\cite{freixas2017}. Applied here, it turns sampler-derived signed
ridges into a quantitative certificate of nonadditivity. The
four-corner pattern is its two-coordinate special case; the theorem
does not assert that every noncoordinate ridge changes a decision.

\resultlead{Exact minimization and realized-margin selection}

\begin{proposition}[Exact minimization of the three-parameter
first-order threat surrogate]
\label{prop:ridge-minimum}
Suppose the rotation--translation response has the form
\begin{equation}
\begin{split}
F(r,t_x,t_y)={}&\ell_r r+\ell_x t_x+\ell_y t_y\\
&+\sum_i c_i|a_i r+t_x|\\
&+\sum_j d_j|b_j r+t_y|,
\end{split}
\label{eq:separated-ridges}
\end{equation}
with signed coefficients $c_i,d_j$ and
$|a_i|,|b_j|<1$. Its global minimum on $[-1,1]^3$
is attained at $r\in\{-1,0,1\}$, with each translation
chosen among $\{-1,1\}$ and its associated ridge
breakpoints $\{-a_i r\}$ or $\{-b_j r\}$, respectively.
Consequently, the global minimum of this first-order
surrogate is computable from finitely many candidates,
without sampling the continuous parameter cube.
\end{proposition}

\begin{proof}
At fixed $r$, the terms involving $t_x$ and $t_y$ separate.
Each is piecewise affine and attains its minimum at an endpoint
or a ridge breakpoint. On either interval $r\in[0,1]$
or $r\in[-1,0]$, all these candidates remain feasible,
and the objective evaluated at each candidate is affine in $r$:
for endpoint candidates the signs of $\pm1+a_i r$
are fixed by $|a_i|<1$, and for breakpoints the
remaining absolute-value terms are proportional to $|r|$.
The minimum over the candidates is therefore concave in $r$
on each half-interval. Adding the two concave minimum-value
functions and $\ell_r r$ preserves concavity; a minimum on
each half-interval is attained at an endpoint.
\end{proof}

This exact search concerns $F$, not the nonlinear network margin.
A nonzero reconstruction defect $d_{\mathrm{ax}}$ does not alone
imply a different minimizing transformation; selection requires a
gap, quantified next.

\begin{proposition}[Selection gap and realized-margin regret]
\label{prop:selection-regret}
Let $\mathcal U=[-1,1]^q$ and define
\begin{equation}
\begin{gathered}
F_{\mathrm{ax}}(U)=\ell^\top U+\sum_{i=1}^q K_i|U_i|,
\quad I(U)=F(U)-F_{\mathrm{ax}}(U),\\
u_{\mathrm{ax}}\in\arg\min_{\mathcal U}F_{\mathrm{ax}},
\qquad u_{\mathrm{joint}}\in\arg\min_{\mathcal U}F,\\
d_{\mathrm{sel}}=F(u_{\mathrm{ax}})-F(u_{\mathrm{joint}})\ge0.
\end{gathered}
\label{eq:selection-gap}
\end{equation}
The axiswise choice is suboptimal for the coupled surrogate
if and only if
\begin{equation}
\exists V\in\mathcal U:\quad
I(u_{\mathrm{ax}})-I(V)>
F_{\mathrm{ax}}(V)-F_{\mathrm{ax}}(u_{\mathrm{ax}}).
\label{eq:selection-criterion}
\end{equation}
Assume the uniform sampling-displacement expansion in
Section~4.2 on $\mathcal U$ and that $\nabla m$ is locally
Lipschitz near $x$. For all sufficiently small $s$, there is
$C<\infty$, independent of $s$, such that, with
$M_s(U)=m(T_{sU}(x))$ and $\gamma_s=\min_{\mathcal U}M_s$,
\begin{equation}
\begin{aligned}
0\le M_s(u_{\mathrm{joint}})-\gamma_s&\le2Cs^2,\\
\big|M_s(u_{\mathrm{ax}})-\gamma_s-sd_{\mathrm{sel}}\big|
&\le2Cs^2,\\
\big|M_s(u_{\mathrm{ax}})-M_s(u_{\mathrm{joint}})
-sd_{\mathrm{sel}}\big|&\le2Cs^2.
\end{aligned}
\label{eq:selection-regret}
\end{equation}
Thus if $d_{\mathrm{sel}}>0$, axiswise selection has
$\Theta(s)$ realized-margin regret, whereas coupled selection
has $O(s^2)$ regret. The statement concerns attained fixed-pair
margins, not distances between minimizers.
\end{proposition}

\begin{proof}
Since $F=F_{\mathrm{ax}}+I$ and $u_{\mathrm{ax}}$ minimizes
$F_{\mathrm{ax}}$, strict improvement by a competitor $V$ is
equivalent to~\eqref{eq:selection-criterion}.
On the finite pixel grid, the expansion of the bilinear
weights in~\eqref{eq:bilinear} and the sampling-displacement
assumption give
$\sup_{\mathcal U}|Z_s-sF|\le C_1s^2$.
They also give $\sup_{\mathcal U}\|\delta_s\|\le C_2s$.
The locally Lipschitz margin gradient yields
$\sup_{\mathcal U}|\Delta_s-Z_s|\le C_3s^2$.
Hence
$\sup_{\mathcal U}|M_s-(m(x)+sF)|\le Cs^2$.
In particular,
$|\gamma_s-[m(x)+s\min_{\mathcal U}F]|\le Cs^2$.
Applying the same bound at the two selected transformations
proves~\eqref{eq:selection-regret}.
\end{proof}

This is a local guarantee for \emph{attained fixed-pair margin}
regret, not multiclass top-1 error or vulnerability ranking. The
separation is strict within the ridge class: on $[-1,1]^2$,
let $F(r,t)=(r-t)/4-|r/2+t|$ and
$F_{\mathrm{ax}}(r,t)=(r-t)/4-|r|/2-|t|$.
The unique axiswise minimizer is $(-1,1)$, with $F=-1$,
whereas $\min F=-3/2$, giving $d_{\mathrm{sel}}=1/2$.
Nonzero interactions can also yield $d_{\mathrm{sel}}=0$
when the minimizers coincide. The regret bound applies in any
finite $q$, whereas Proposition~\ref{prop:ridge-minimum}
uses the special three-parameter geometry for exact minimization.

Define the worst-case first-order margin
$\widehat\gamma_s=m(x)+s\min_{U\in[-1,1]^3}F(U)$.
If a uniform remainder estimate
$\sup_U|\Delta_s(U)-sF(U)|\le\varepsilon_s$ is available,
then
\begin{equation}
\widehat\gamma_s-\varepsilon_s
\le\min_U m(T_{sU}(x))
\le\widehat\gamma_s+\varepsilon_s.
\label{eq:margin-threat-bound}
\end{equation}
Thus $\widehat\gamma_s>\varepsilon_s$ certifies preservation
of the \emph{selected pairwise margin} on the box, while
$\widehat\gamma_s<-\varepsilon_s$ ensures a pairwise crossing.
These statements are conditional and fixed-pair: a positive
bound for one margin cannot certify multiclass robustness.
No calibrated $\varepsilon_s$ is available at the tested scales.

\resultlead{Multiclass top-1 event prediction}
The same sampler geometry can be reused for several logit margins.
For the clean top-1 class $k$, write
$m_j(z)=f_k(z)-f_j(z)$ for $j\ne k$, and let
$\mathcal J\subseteq\{1,\dots,K\}\setminus\{k\}$
be a fixed set of competitors selected \emph{before} transforming the
image. Each clean gradient $g_j=\nabla m_j(x)$ yields the first-order
ridge response $F_j(U)$ of Section~4.2.
We define the clean-gradient multiclass approximation
\begin{equation}
\begin{split}
\widehat b_{\mathcal J,s}(U)
&=\min_{j\in\mathcal J}\{m_j(x)+sF_j(U)\},\\
\widehat y_s(U)
&=\mathbf 1\{\widehat b_{\mathcal J,s}(U)<0\}.
\end{split}
\label{eq:multiclass-event}
\end{equation}
Equation~\eqref{eq:multiclass-event} predicts whether the original
top-1 class changes at a specified $U$, not the replacement-class
identity. Unlike the fixed-pair constructions above, any included
competitor can trigger this binary event. Axiswise and
finite-difference local responses are compared on the same frozen
directions in Section~\ref{sec:exp-multiclass}.

The gap between a predicted event and the true all-class decision
has two sources: approximation error on the included margins and
new winners omitted by the clean-image shortlist. Set
$b_{\mathcal J,s}(U)=\min_{j\in\mathcal J}m_j(T_{sU}(x))$
and suppose
\begin{equation}
\max_{j\in\mathcal J}\sup_{U\in\mathcal U}
|m_j(T_{sU}(x))-m_j(x)-sF_j(U)|\le\varepsilon_s.
\label{eq:multi-remainder}
\end{equation}
The minimum is 1-Lipschitz in the maximum norm, hence
\begin{equation}
|b_{\mathcal J,s}(U)-\widehat b_{\mathcal J,s}(U)|
\le\varepsilon_s.
\label{eq:multi-bound}
\end{equation}
Let $y_s(U)=\mathbf 1\{\min_{j\ne k}m_j(T_{sU}(x))<0\}$
be the actual all-class top-1 event. Then
\begin{equation}
\begin{split}
\{\widehat y_s\ne y_s\}\ \subseteq\
&\{\,|\widehat b_{\mathcal J,s}|\le\varepsilon_s\,\}\\
&\cup\{\,b_{\mathcal J,s}\ge0,
\min_{j\notin\mathcal J,\,j\ne k}
 m_j(T_{sU}(x))<0\,\}.
\end{split}
\label{eq:multi-disagreement}
\end{equation}
The omitted-class set vanishes with all competitors included.
Locally Lipschitz margin gradients and a uniform sampler expansion
yield an $O(s^2)$ remainder in~\eqref{eq:multi-remainder}, but
its constant is not calibrated experimentally. These standard
minimum/sign-stability implications neither certify the network
nor order the top-1 errors of coupled, axiswise, and finite-difference
models; Section~\ref{sec:exp-multiclass} compares them directly.

\begin{corollary}[Exact multiclass surrogate selection and conditional regret]
\label{cor:multiclass-selection}
Let $\mathcal U=[-1,1]^3$ and let $\mathcal J$ be a finite,
nonempty set of clean-image competitors. Suppose each $F_j$ has the
rotation--translation ridge form in Proposition~\ref{prop:ridge-minimum}
with $|a_i|,|b_i|<1$. Define
\[
\begin{aligned}
 b_{\mathcal J,s}(u)&=\min_{j\in\mathcal J}m_j(T_{su}(x)),\\
 \widehat b_{\mathcal J,s}(u)&=\min_{j\in\mathcal J}
           \{m_j(x)+sF_j(u)\},\\
 \widehat b_{\mathrm{ax},\mathcal J,s}(u)&=
           \min_{j\in\mathcal J}\{m_j(x)+sF_{\mathrm{ax},j}(u)\}.
\end{aligned}
\]
For $s>0$, the surrogate minimum is computable exactly by
\[
 \min_{u\in\mathcal U}\widehat b_{\mathcal J,s}(u)
 =\min_{j\in\mathcal J}\left\{m_j(x)+s\min_{u\in\mathcal U}F_j(u)\right\},
\]
where each inner minimum is given by the finite candidate set in
Proposition~\ref{prop:ridge-minimum}. Let $u_{\rm joint}$ and
$u_{\rm ax}$ minimize $\widehat b_{\mathcal J,s}$ and
$\widehat b_{\mathrm{ax},\mathcal J,s}$, respectively, and set
\[
\begin{aligned}
\gamma_{\mathcal J,s}&=\min_{u\in\mathcal U}b_{\mathcal J,s}(u),\\
G_{\mathcal J,s}&=\widehat b_{\mathcal J,s}(u_{\rm ax})
 -\widehat b_{\mathcal J,s}(u_{\rm joint})\ge0.
\end{aligned}
\]
If
\[
\sup_{u\in\mathcal U}|b_{\mathcal J,s}(u)
          -\widehat b_{\mathcal J,s}(u)|\le\varepsilon_s,
\]
then
\begin{equation}
\begin{aligned}
0\le b_{\mathcal J,s}(u_{\rm joint})-\gamma_{\mathcal J,s}
    &\le2\varepsilon_s,\\
\big|b_{\mathcal J,s}(u_{\rm ax})-\gamma_{\mathcal J,s}
         -G_{\mathcal J,s}\big|&\le2\varepsilon_s,\\
\big|b_{\mathcal J,s}(u_{\rm ax})-b_{\mathcal J,s}(u_{\rm joint})
         -G_{\mathcal J,s}\big|&\le2\varepsilon_s.
\end{aligned}
\label{eq:multi-select-regret}
\end{equation}
In particular, if $\varepsilon_s=O(s^2)$, the coupled selection
has $O(s^2)$ regret for the selected-class minimum. If moreover
$G_{\mathcal J,s}/s$ is bounded away from zero as $s\downarrow0$,
axiswise selection has $\Theta(s)$ regret.
When $\mathcal J$ contains every competitor $j\ne k$, the statements
concern the all-class top-1 margin. In that case,
$\min_u\widehat b_{\mathcal J,s}(u)>\varepsilon_s$ excludes
strict top-1 changes on $\mathcal U$, whereas
$\min_u\widehat b_{\mathcal J,s}(u)<-\varepsilon_s$ ensures a
change at $u_{\rm joint}$.
\end{corollary}

\begin{proof}
The order of the two minima can be exchanged because
$\mathcal J$ is finite. Proposition~\ref{prop:ridge-minimum}
computes each $\min_u F_j(u)$ exactly. The pointwise bound implies
$|\gamma_{\mathcal J,s}-\min_u\widehat b_{\mathcal J,s}(u)|
\le\varepsilon_s$. It also applies at $u_{\rm joint}$ and
$u_{\rm ax}$. Combining these estimates gives the three
inequalities in~\eqref{eq:multi-select-regret}. The asymptotic
statements follow from the error orders, nonnegative regret,
and $G_{\mathcal J,s}=O(s)$, since the finite family of ridge responses
is bounded on the compact parameter cube. For the all-class case, a strictly positive minimum margin
preserves the clean top-1 class, while a strictly negative margin
implies a different top-1 class.
\end{proof}

The argument combines the standard stability of finite minima with
our sampler-specific exact ridge minimization. For a fixed image
with a strictly unique closest clean competitor, that competitor
also minimizes the realized and surrogate margins uniformly for
all sufficiently small $s$, so the all-class conclusion then
reduces locally to the fixed-pair result. For a restricted shortlist,
\eqref{eq:multi-select-regret} concerns only its included margins;
additional control of omitted competitors is required for
all-class claims. No uniform numerical remainder constant is
calibrated at the experimental scales.

\section{Experimental validation}
\label{sec:experiments}

\subsection{Evaluation design}
\label{sec:eval-design}
We test three distinct consequences of the theory: fidelity of the clean-gradient projection, transfer to the \emph{realized fixed-pair} network margin, and decision-level behavior. CIFAR-10/ResNet-18 experiments use bilinear resampling with border padding and \texttt{align\_corners=False} \cite{krizhevsky2009,he2016}. All ImageNetV2/ConvNeXt-Tiny evaluations use the frozen bilinear \texttt{grid\_sample} operator with border padding, \texttt{align\_corners=True}, and no output clipping \cite{recht2019imagenetv2,liu2022convnext}. Images are bilinearly resized to a shortest side of 236 pixels (with antialiasing), center-cropped to $224\times224$, transformed in raw $[0,1]$ coordinates, and ImageNet-normalized before network inference. For the ImageNetV2 protocol just described, the rotation angle is $sU_{\rm rot}/R$ radians and the translations are $(sU_x,sU_y)$ pixels, where $R$ is the half-diagonal of the pixel grid. The independent CIFAR-10 threat-search experiment uses the distinct base rotation and translation scales stated in Sec.~\ref{sec:exp-decision}. Class pairs and low-margin cohorts are selected before transformation evaluations; independent confirmations exclude pilot images. We report root-mean-square (RMS) reconstruction error, fixed-pair crossing-probability mean absolute error (MAE), top-1 screening, and the actual margin attained by a method-selected transformation as \emph{different endpoints}. Confidence intervals resample images (stratified by cohort where indicated), not transformation directions. Matched and direction-held-out analyses that reuse images are distinguished from new-image confirmations.

\resultlead{Independent multiclass confirmation}
An independent study evaluates TorchVision
ConvNeXt-Tiny (ImageNet-1K pretrained) on ImageNetV2
MatchedFrequency, with the same ImageNet preprocessing and sampler
conventions as above but on newly selected images. It excludes
1,002 previously screened image identities from earlier external
experiments and freezes pilot/confirmation pools, clean-image
competitors, and antithetic directions before evaluating the
transformed network. The main binary top-1 endpoint is evaluated
on a 256-image confirmation cohort at the prespecified
$s=.40$; $s=.10$ is secondary. This experiment does not reuse the
images of the earlier ImageNetV2 projected or diagnostic evaluations.

\subsection{First-order incompleteness and coupled reconstruction}
\label{sec:exp-reconstruction}
\begin{figure*}[!t]
\centering
\includegraphics[width=.93\textwidth]{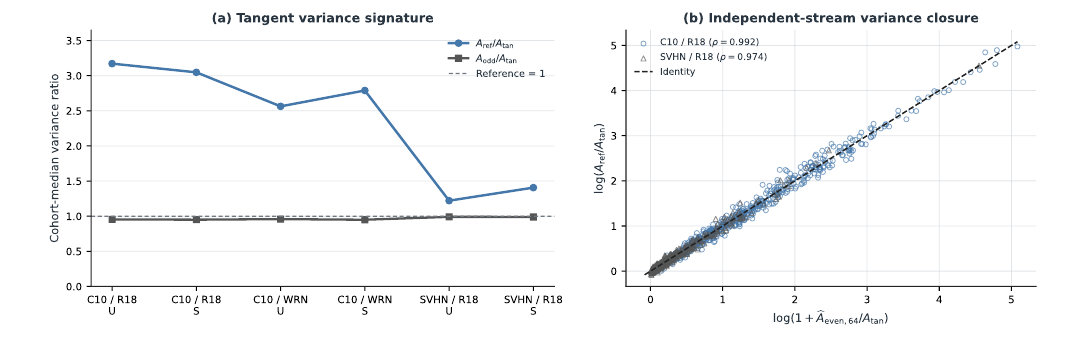}
\caption{Observed tangent-variance defect across model and cohort controls. (a) Realized and odd projected variances relative to the central-tangent variance across CIFAR-10/ResNet-18, CIFAR-10/WRN \cite{zagoruyko2016} and SVHN/ResNet-18 \cite{netzer2011} (uniform and low-margin cohorts). (b) Even-response variance accounts for the measured variance gap using independent estimation streams. This identity does not itself establish odd-response matching.}
\label{fig:core}
\end{figure*}

Across CIFAR-10/ResNet-18, CIFAR-10/WRN and SVHN/ResNet-18, the measured even projected component accounts for the variance defect of the central tangent (Fig.~\ref{fig:core}). The exact three-pixel construction (Fig.~\ref{fig:toy}) additionally gives different crossing probabilities for responses sharing the same parameter-linear tangent.

\begin{figure*}[!t]
\centering
\includegraphics[width=.84\textwidth]{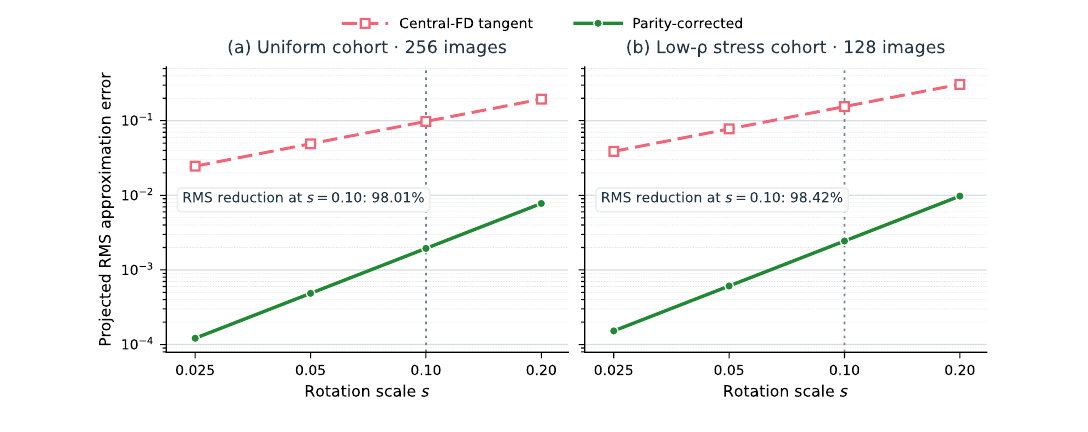}
\caption{Independent CIFAR-10 scalar-rotation confirmation: projected RMS of the central tangent and the analytic parity correction across transformation scales. The vertical marker denotes the prespecified $s=.10$.}
\label{fig:scalar}
\end{figure*}

On 256 new uniform and 128 new low-margin CIFAR-10 images at $s=.10$, the scalar parity correction reduces projected RMS from $.09786$ to $.00194$ and from $.15468$ to $.00244$, respectively (Fig.~\ref{fig:scalar}). The corresponding realized fixed-pair RMS values are $.10037\to.01339$ and $.15750\to.01555$. The odd-only control agrees with the central tangent; the measured scale dependence is consistent with the first- versus second-order errors predicted by Eq.~\eqref{eq:scalar}.

\begin{figure*}[!t]
\centering
\includegraphics[width=.92\textwidth]{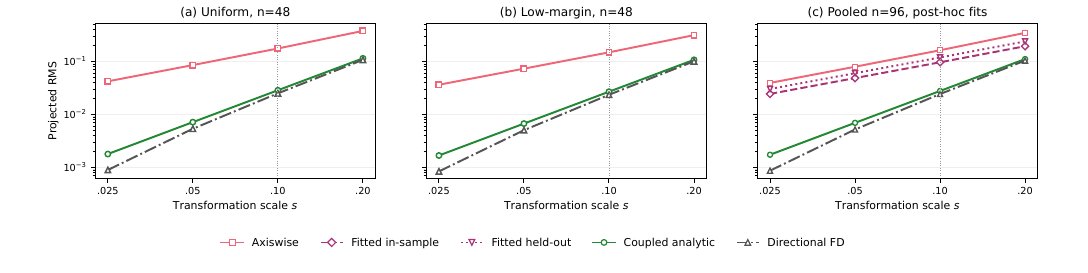}
\caption{Joint rotation--translation responses on independent CIFAR-10 uniform and low-margin cohorts (a,b). Panel (c) compares coordinate-affine--absolute fits evaluated in sample and on held-out directions of these same images.}
\label{fig:coupled}
\end{figure*}

The principal joint-transformation confirmation uses 48 uniform and 48 low-margin CIFAR-10 images, disjoint from the preceding scalar study. At $s=.10$, axiswise assembly has pooled projected RMS $.16377$, versus $.02791$ for the coupled response: an \emph{82.96\% reduction} (paired image-bootstrap 95\% interval $[80.89,84.81]\%$). The observed RMS slopes are approximately one and two, respectively (Fig.~\ref{fig:coupled}a,b). One-sided directional differences reach $.02442$. On the same 96 images, a retrospective coordinate-affine--absolute fit gives RMS $.09708$ in sample and $.11998$ on held-out directions, versus $.02791$ for the coupled formula (Fig.~\ref{fig:coupled}c). This fitted comparison tests the finite-dimensional class of Proposition~\ref{prop:additive}, rather than unrestricted additive models.

\begin{figure*}[!t]
\centering
\includegraphics[width=.92\textwidth]{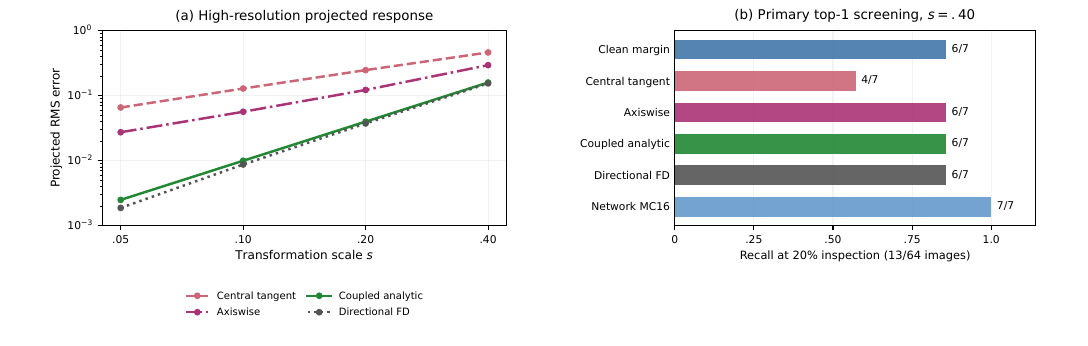}
\caption{ImageNetV2/ConvNeXt-Tiny evaluations. (a) Independent projected reconstruction on 64 uniform images. (b) Prospective top-1 screening on a separate 64-image uniform cohort at $s=.40$: recall among seven positives at an inspection budget of 13 images.}
\label{fig:external}
\end{figure*}

An independent ImageNetV2/ConvNeXt-Tiny projected confirmation on 64 uniform images gives axiswise, coupled, and directional RMS values $.05653$, $.00996$, and $.00872$, respectively, at $s=.10$ (Fig.~\ref{fig:external}a): the coupled reduction is \emph{82.38\%}. Table~\ref{tab:reconstruction} distinguishes this new-image projected confirmation from the separate, image-reusing realized-margin evaluations.

\begin{table*}[!t]
\centering
\caption{Reconstruction RMS at $s=.10$ for the projected and realized fixed-pair responses. The CIFAR-10 realized evaluation reuses the 96-image joint cohort; the ImageNetV2 realized evaluation uses newly held-out directions on 128 previously selected images and is not another new-image confirmation. FD denotes one-sided directional differences.}
\label{tab:reconstruction}
\footnotesize
\setlength{\tabcolsep}{5pt}
\renewcommand{\arraystretch}{1.15}
\begin{tabular}{@{}llrrrr@{}}
\toprule
Dataset / network & Response & Images & Axiswise & Coupled & FD\\
\midrule
CIFAR-10 / ResNet-18 & Projected & 96 & $.16377$ & $.02791$ & $.02442$\\
CIFAR-10 / ResNet-18 & Realized fixed-pair & 96 & $.19456$ & $.11723$ & $.11665$\\
ImageNetV2 / ConvNeXt-Tiny & Projected & 64 & $.05653$ & $.00996$ & $.00872$\\
ImageNetV2 / ConvNeXt-Tiny & Realized fixed-pair & 128 & $.08734$ & $.05002$ & --\\
\bottomrule
\end{tabular}
\end{table*}

\subsection{Transfer to realized margins and sampler controls}
\label{sec:exp-realized}
For a fixed clean class pair, write $\Delta_s(U)=m(T_{sU}(x))-m(x)$. The correction residual separates exactly into
\begin{align}
\Delta_s-\widehat Z_{\mathrm{joint},s}
&=\underbrace{\Delta_s-Z_s}_{e_{\rm clf}}
 +\underbrace{Z_s-\widehat Z_{\mathrm{joint},s}}_{e_{\rm samp}},
\label{eq:errors}
\end{align}
where the two terms arise from classifier linearization and sampler approximation, respectively.

\begin{figure*}[!t]
\centering
\includegraphics[width=.90\textwidth]{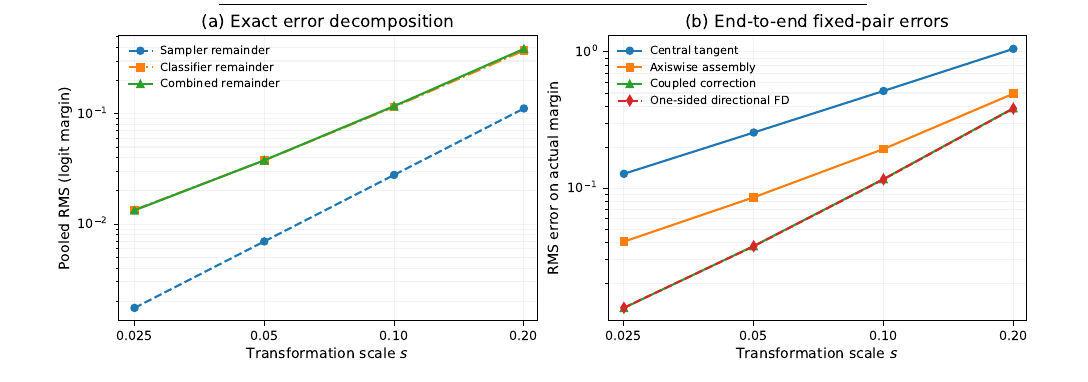}
\caption{Matched realized fixed-pair margin evaluation on 96 CIFAR-10/ResNet-18 images and frozen joint directions. (a) Classifier and sampler residuals. (b) RMS reconstruction errors across the four evaluated scales.}
\label{fig:realmargin}
\end{figure*}

Using the same 96 CIFAR-10 images and frozen transformation directions as the joint projected confirmation, the realized-margin RMS decreases from $.19456$ (axiswise) to $.11723$ (coupled), a \emph{39.75\% reduction} (stratified paired image-bootstrap 95\% interval $[34.53,44.50]\%$; Fig.~\ref{fig:realmargin}). The central tangent and one-sided directional differences give $.51761$ and $.11665$. At $s=.10$, the classifier and sampler residuals in Eq.~\eqref{eq:errors} have RMS $.11546$ and $.02791$: classifier nonlinearities dominate the corrected error. The fitted finite-scale slopes are $1.20$ for axiswise error, $2.00$ for the sampler residual, and $1.62$ for total coupled error.

The effect is parity-selective. At $s=.10$, the odd RMS component of realized reconstruction error is $.05888$ for both analytic surrogates, while the even component decreases from $.18543$ to $.10137$; mean-response drift error falls from $.12663$ to $.07963$. On ImageNetV2/ConvNeXt-Tiny, 128 previously selected images evaluated on new held-out direction pairs exhibit realized RMS $.08734\to.05002$ (\emph{42.73\%}; exploratory stratified 95\% interval $[33.12,52.68]\%$), with the same unchanged-odd/reduced-even pattern. This external result involves a distinct evaluation protocol from the independent 64-image projected confirmation.

Additional sampler controls localize the mechanism: unclipped affine brightness--contrast has algebraically zero projected even response, while clipping generates a nonzero even term; unclipped bicubic sampling exhibits approximately second-order rather than first-order even scaling. These supplementary mechanistic checks are descriptive and not independent confirmations.

\subsection{Interaction diagnostics and decision-event estimates}
\label{sec:exp-diagnostics}
\begin{figure*}[!t]
\centering
\includegraphics[width=.92\textwidth]{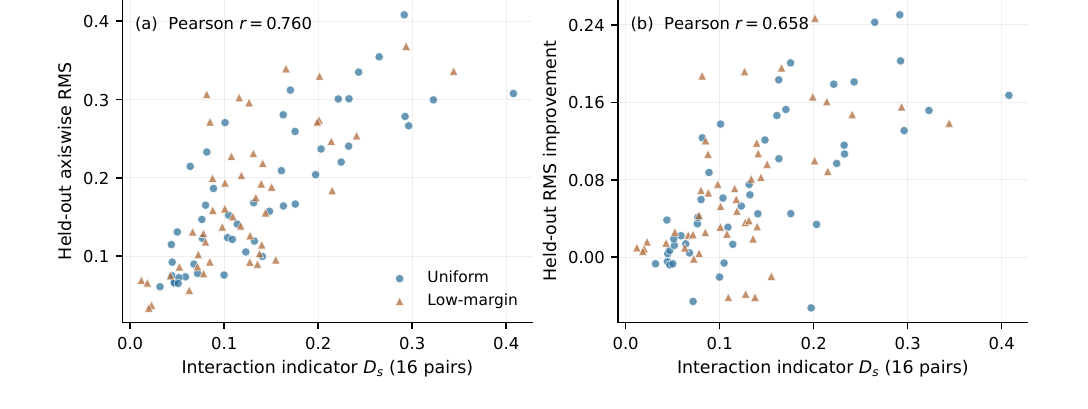}
\caption{CIFAR-10 direction-held-out interaction diagnostic ($n=96$, $s=.10$). The empirical score $\widehat D_{s,16}$ uses 16 antithetic pairs per image; 16 distinct pairs determine (a) realized axiswise RMS and (b) realized absolute RMS improvement.}
\label{fig:diagnostic}
\end{figure*}

The analytic interaction indicator $D_s=s\,d_{\mathrm{ax}}$ measures the part of the first-order response omitted by axiswise assembly, without transformed-network inference. To test its relation to \emph{realized} approximation error, the exploratory CIFAR-10 analysis separates 16 antithetic scoring pairs from 16 held-out evaluation pairs on each of the same 96 images. The reported correlations use the empirical score $\widehat D_{s,16}(x)=[16^{-1}\sum_{i=1}^{16}I_s(U_i)^2]^{1/2}$, evaluated from the analytic interaction $I_s=\widehat Z_{\mathrm{joint},s}-\widehat Z_{\mathrm{ax},s}$ on scoring directions only; the antithetic values are identical because $I_s$ is even. At $s=.10$, $\widehat D_{s,16}$ correlates with realized axiswise RMS at $r=.760$ (stratified image-bootstrap 95\% interval $[.670,.837]$), versus $r=.534$ for the full-even amplitude $Q_s=\|sH\|$ (Fig.~\ref{fig:diagnostic}). The correlation with realized absolute correction gain is $.658$. These are direction-held-out associations on previously selected images.

A second, protocol-locked direction-held-out evaluation uses 128 previously selected ImageNetV2 images (64 uniform, 64 low-margin) and ConvNeXt-Tiny with \texttt{align\_corners=True}. On 16 scoring and 16 disjoint evaluation antithetic pairs per image, $\widehat D_{s,16}$ correlates with realized axiswise error at $r=.722$, versus $.704$ for $Q_s$ and $.720$ for the axiswise-even amplitude $B_s$. The prespecified $D_s$--$Q_s$ difference is only $.018$ (exploratory bootstrap interval $[-.096,.136]$); \emph{superiority is not established} for the primary diagnostic endpoint. The secondary realized-gain correlations are $.774$, $.620$, and $.672$, respectively. These secondary correlations are exploratory, and all three scores are evaluated on the same independently held-out direction pairs.

Decision-event evaluation yields a more differentiated picture. On 128 new low-margin CIFAR-10 images at the pilot-informed $s=.40$, the fixed-pair crossing-probability MAE improves from $.07581$ (central tangent) to $.02277$ (analytic correction), compared with $.02219$ for directional differences and $.01267$ for direct-network MC64. In an independent prospective ImageNetV2 top-1 screening experiment (Fig.~\ref{fig:external}b; $n=64$, $s=.40$), both the coupled score and clean margin retrieve six of seven positives within an inspection budget of 13 images; direct-network MC16 retrieves seven. Thus gains in local reconstruction and fixed-pair crossing estimation do not imply a general advantage in ranking vulnerable images.

\subsection{Decision consequences: nonadditivity and threat search}
\label{sec:exp-decision}
\begin{figure*}[!tbp]
\centering
\includegraphics[width=.98\textwidth]{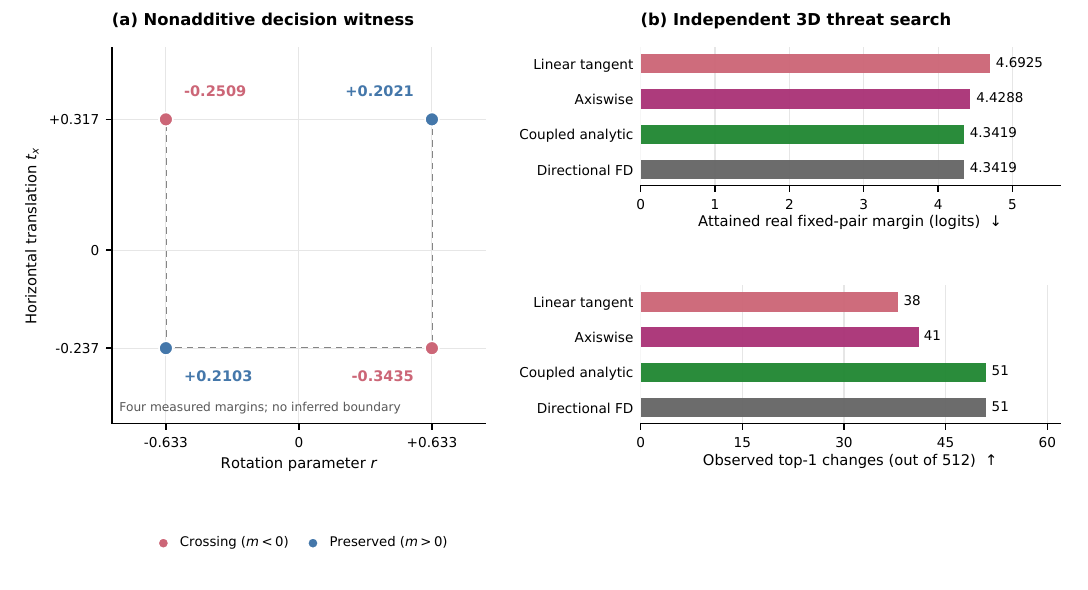}
\caption{Decision consequences of coupled geometric transformations.
(a) Realized fixed-pair margins at four prospectively selected
rotation--translation settings (CIFAR-10 image 147, $s=.10$).
Markers denote network evaluations; dashed segments do not
represent a decision boundary. The checkerboard sign pattern
is incompatible with any coordinate-additive margin.
(b) Independent threat search on 512 CIFAR-10 images:
mean realized fixed-pair margins and observed top-1 changes
for transformations selected by each method.}
\label{fig:decision-consequences}
\end{figure*}

\resultlead{Prospectively selected nonadditive witnesses}
The frozen CIFAR-10/ResNet-18 protocol selects 512 previously unexamined images (256 uniform and 256 low-margin; $s=.10$). Ridge coefficients computed from each clean image and margin gradient propose four joint rotation--translation settings \emph{before} transformed-network evaluation. Of 27 proposed checkerboards, 10 retain the alternating fixed-pair signs on realized logits, and nine meet the prespecified minimum absolute margin of $.005$ logit. The strict witness for test image 147 has realized margins $(-.2509,.2103,.2021,-.3435)$ in row order (Fig.~\ref{fig:decision-consequences}a). These four-point signs rule out any coordinate-additive representation of that fixed-pair decision pattern; one-sided directional differences also recover all ten observed sign patterns. The equal-marginal certificate of Proposition~\ref{prop:finite-additive-sign}, applied \emph{retrospectively} to these four measured margins, yields a lower bound of $.2517$ logit on the maximum error of \emph{any} coordinate-additive fit to the observed values. This calculation is not an independent test of the general $q$-parameter criterion.

\resultlead{Independent continuous 3D threat search}
A \emph{separate} 512-image CIFAR-10/ResNet-18 cohort, disjoint from all previous pools, tests whether minimizing the analytic coupled surrogate finds lower \emph{realized} network margins than minimizing the axiswise surrogate. Each method selects a direction on $[-1,1]^3$ at $s=.10$ (base rotation $2^\circ$, translations one pixel), before the network evaluates the selected transform. For the $32\times32$ grid, the geometric coefficient bound $15.5(2\pi/180)\simeq.5411<1$ ensures applicability of Proposition~\ref{prop:ridge-minimum}: optimization is global for each first-order surrogate.

The coupled-selected transformation attains mean realized margin $4.3419$, versus $4.4288$ for axiswise assembly and $4.6925$ for the linear tangent (Fig.~\ref{fig:decision-consequences}b). The prespecified paired gain against axiswise assembly is $.08689$ logit (stratified image-bootstrap 95\% interval $[.07290,.10147]$; $10{,}000$ resamples). The coupled search identifies 51 top-1 changes versus 41 for axiswise assembly (11 coupled-only and one axis-only; secondary exact paired McNemar $p=.00635$). The realized margin is lower for coupled selection on 279 images, higher on 49, and identical on 184 images where the methods select the same transformation. This finite-scale paired comparison is consistent with the selection effect in Proposition~\ref{prop:selection-regret}, but does not estimate $d_{\mathrm{sel}}$ imagewise or test the asymptotic regret rates. A directional finite-difference search attains mean margin $4.34194$ and 51 changes. Its $3\times17\times17$ grid is augmented with the three analytic minimizers, \emph{including the coupled one}; the resulting comparison is therefore not independent in candidate support or matched in sampler-evaluation cost. Network logits are used only to assess the selected transformations: the global optimum belongs to the first-order surrogate, not to the nonlinear classifier margin.

\begin{figure*}[!t]
\centering
\includegraphics[width=.77\textwidth]{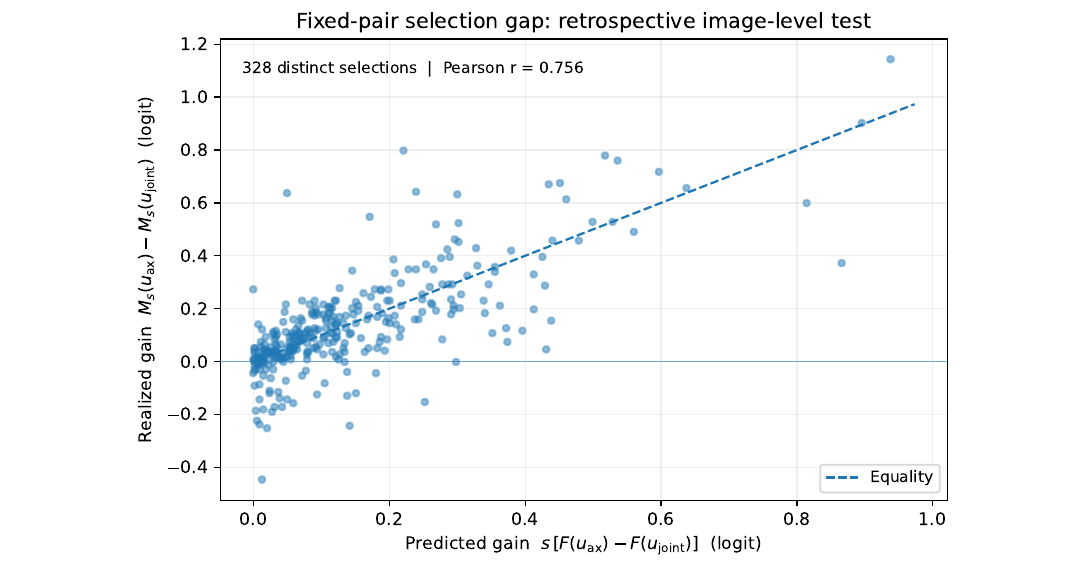}
\caption{Retrospective fixed-pair selection-gap diagnostic at
$s=.10$ on 328 CIFAR-10/ResNet-18 images for which axiswise and
coupled methods selected different transformations. The horizontal
axis is the coupled first-order predicted gain; the vertical axis
is the realized network-margin gain. The dashed diagonal denotes
equality. The image pool and transformation choices were fixed by
the prior 512-image threat-search experiment; this is not an
independent confirmation or a multiclass top-1 test.}
\label{fig:selection-gap-imagelevel}
\end{figure*}

\resultlead{Retrospective validation of the selection-gap mechanism}
To examine the mechanism in Proposition~\ref{prop:selection-regret}
more directly, we recomputed the coupled first-order response at
both transformations frozen in the independent 512-image CIFAR-10
fixed-pair search. For each image, the predicted gain is
$P_s=s[F(u_{\mathrm{ax}})-F(u_{\mathrm{joint}})]$, while the
realized gain is $A_s=M_s(u_{\mathrm{ax}})-M_s(u_{\mathrm{joint}})$.
At the original $s=.10$, mean $P_s$ is $.09153$ logit and mean
$A_s$ is $.08689$ logit. Among the 328 images with different
selected transformations, their Pearson correlation is $.756$
(retrospective stratified image-bootstrap 95\% interval
$[.676,.825]$); the 184 identical selections are omitted from
this correlation to avoid inflating it with zero--zero pairs
(Fig.~\ref{fig:selection-gap-imagelevel}). On 49 of those
328 images, the realized ordering favors axiswise selection.
This post hoc comparison supports an association between the
predicted gap and the realized gain but neither measures the
true nonlinear-network optimum nor establishes the asymptotic
regret rate.

\subsection{Independent multiclass top-1 decision confirmation}
\label{sec:exp-multiclass}
\resultlead{Registered design and comparison}
We test whether the recovered sampler interactions improve the
\emph{binary prediction of an actual top-1 class change} on a
separate architecture. ImageNetV2/ConvNeXt-Tiny (ImageNet-1K
weights) supplies a 64-image pilot and an independently sampled
256-image confirmation, with 32 and 128 images per stratum,
respectively. One stratum is uniformly selected from a frozen
screening pool and the other consists of the lowest clean-margin
images among a disjoint set of screening candidates. The
confirmation excludes 1,002 earlier ImageNetV2 screening identities;
all 320 selected images are distinct. Therefore the pooled 50/50
stratified results are \emph{not} population prevalence estimates
for ImageNetV2.

Each image is evaluated on 64 frozen antithetic joint
rotation--translation directions. For the original clean winner
$k$, the 20 closest clean competitors are fixed before any
transformed inference; actual top-1 changes are established using
\emph{all 1,000 logits} at the transformed images. The analytic
models use the same 20 margins: a clean-gradient axiswise
reconstruction and the corresponding signed-ridge coupled
reconstruction. A coordinate finite-difference (FD) baseline
uses the six signed coordinate displacements at
$h=.0125$ to form a coordinatewise odd/even response model.
A directional FD baseline evaluates the actual network on each of
the same 64 directions, but at the \emph{small step} $h=.0125$,
then extrapolates to the target amplitude. Its evaluations are
not the target network labels. The registration fixes the primary
comparison as axiswise versus coupled binary top-1 error on the
256 confirmation images at $s=.40$. The separate $s=.10$
results and pilot are secondary.

\begin{table*}[!t]
\centering
\caption{Independent ImageNetV2/ConvNeXt-Tiny multiclass top-1
confirmation (256 images; 64 common transforms per image and scale;
16,384 binary decisions per scale). Entries give erroneous flip
predictions, with the corresponding rate. False alarms and missed
flips are shown at the prespecified primary $s=.40$. Cost is the
median recorded model-construction time per image in the supplied
GPU run, not a general matched-runtime benchmark.}
\label{tab:m19-top1}
\footnotesize
\setlength{\tabcolsep}{5pt}
\renewcommand{\arraystretch}{1.14}
\begin{tabular}{@{}lrrrrr@{}}
\toprule
Method & $s=.10$ errors (\%) & $s=.40$ errors (\%) &
False alarms & Missed flips & Time (s/image)\\
\midrule
Axiswise analytic & 427 (2.61) & 1,961 (11.97) & 1,313 & 648 & 0.137\\
\textbf{Coupled ridges} & \textbf{234 (1.43)} &
\textbf{1,676 (10.23)} & 988 & 688 & 0.137\\
Coordinate FD & 408 (2.49) & 1,919 (11.71) & 1,285 & 634 & 0.027\\
Directional FD & 206 (1.26) & 1,639 (10.00) & 966 & 673 & 0.274\\
\bottomrule
\end{tabular}
\end{table*}

\begin{figure*}[!t]
\centering
\includegraphics[width=.91\textwidth]{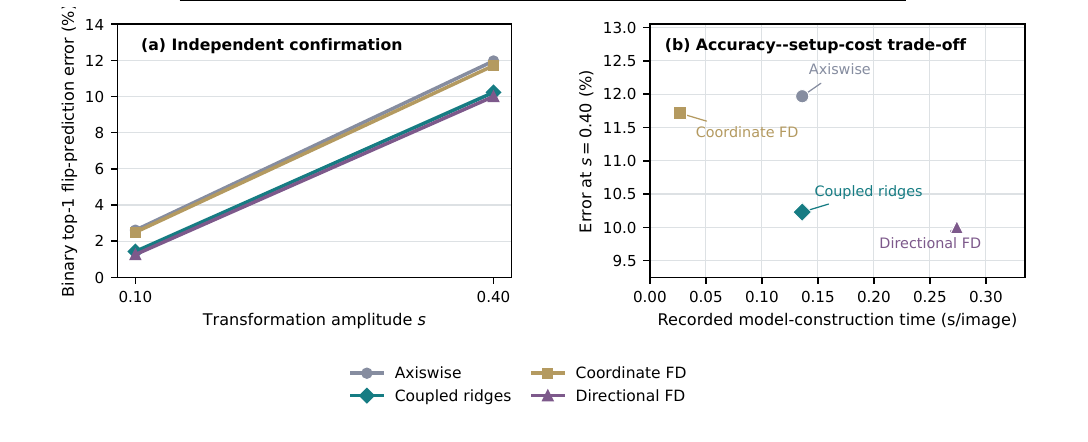}
\caption{Independent multiclass decision confirmation on
ImageNetV2/ConvNeXt-Tiny. (a) Binary top-1 flip-prediction error
across the two registered amplitudes, on the same 256 confirmation
images and 64 directions per image. (b) Primary $s=.40$ error
against median recorded \emph{model-construction} time per image:
the axiswise and coupled analytic predictions share the same
20 clean-gradient and ridge-coefficient computation. The
plot compares the observed implementation costs, not fully
matched end-to-end runtime.}
\label{fig:m19}
\end{figure*}

\resultlead{Primary decision result}
At $s=.40$, the network changes its top-1 prediction on 3,064
of the 16,384 evaluated transformations. Axiswise assembly
misclassifies 1,961 binary events (11.97\%), while the coupled
ridge model misclassifies 1,676 (10.23\%); the relative reduction
in error count is 14.53\%. The \emph{prespecified paired} difference
in error rates is $1.7395$ percentage points (stratified
image-bootstrap 95\% percentile interval $[0.9277,2.6062]$;
20,000 resamples). Improvement occurs within both cohorts:
166 fewer errors among the 8,192 low-margin directions and 119
fewer among the 8,192 uniformly screened directions. At the
secondary $s=.10$, the error counts are 427 and 234,
respectively, giving an improvement of $1.1780$ points
(95\% interval $[0.6775,1.7212]$). Thus a decision-level gain
over independent axiswise assembly persists beyond the
CIFAR-10/ResNet-18 experiments.

\resultlead{Finite differences, misses, and coverage}
The directional FD baseline is slightly \emph{more accurate}
than coupled ridges at the primary scale: 1,639 versus 1,676
errors, a difference of $0.2258$ point in favor of directional FD
(95\% interval $[0.0916,0.3662]$). The coordinate FD baseline,
by contrast, produces 1,919 errors. The coupled reduction relative
to axiswise comprises 325 fewer false alarms and \emph{40 more}
missed changes (Table~\ref{tab:m19-top1}); it is not a uniform
improvement in flip recall. On 15 primary-scale transformations
from one image, the actual new winner lies outside the frozen
20-class shortlist. These events remain in the ground-truth
binary labels and denominator; being outside the shortlist is
not automatically a binary prediction error if another included
margin predicts a crossing.

The median recorded costs for building the 20 analytic margins
and ridge coefficients jointly, six coordinate FD queries, and
64 directional FD queries are $0.1369$, $0.0272$, and $0.2741$
seconds per image, respectively. The analytic model can be
evaluated on further directions without additional network
forwards. The recorded times exclude a separately isolated
measurement of analytic direction evaluation and do not establish
hardware-independent or strict end-to-end cost superiority.
The analytic correction offers a distinct accuracy--construction-
cost trade-off, rather than a claim to dominate directional FD.

\resultlead{Independent and secondary CIFAR-10 multiclass checks}
A prospective 256-image fixed-pair sign study observes 135
axiswise versus 97 coupled sign-prediction errors on 16,384
directions at $s=.10$. Reanalysis of the \emph{same directions}
with multiclass top-1 labels gives 138 versus 106 errors;
this is not a new-image confirmation. A further disjoint
64-image prospective multiclass study (2,048 directions) gives
20 versus 14 top-1 errors at $s=.10$ and 56 versus 39 at
$s=.20$; the $s=.10$ paired image-bootstrap interval touches
zero. These secondary checks support the mechanism but are not
pooled with the external ImageNetV2 confirmation because their image
selection, transformation, and architecture protocols differ.

\section{Discussion}

\resultlead{First-order information beyond coordinate tangents}
For identity-aligned bilinear resampling, the implemented
joint image transformation can carry a first-order response that
is absent from individually assembled coordinate responses.
After projection onto a selected classifier margin, this
interaction has an explicit signed-ridge representation, with
cancellation determined by the image and the margin gradient.
The resulting coupled surrogate attains $O(s^2)$ projected RMS
error, whereas incomplete axiswise assembly has a $\Theta(s)$
defect. Under independent parameter laws, a surviving
noncoordinate ridge also obstructs every square-integrable
coordinate-additive approximation. These statements identify
information that coordinatewise refitting cannot restore.

\resultlead{Consequences for decisions}
The same ridge structure permits prospective construction of
nonadditive fixed-pair sign patterns and exact minimization of
the studied three-parameter first-order threat surrogate.
Conditional local bounds relate coupled selection to realized
fixed-pair margin regret. Experiments support the distinction
between margin reconstruction and decisions: coupled selection
finds more top-1 changes than axiswise selection in an independent
CIFAR-10 search (51 versus 41), while independent
ImageNetV2/ConvNeXt-Tiny evaluation reduces binary multiclass
flip-prediction error from $11.97\%$ to $10.23\%$. Directional
finite differences reach $10.00\%$ on the latter endpoint, at
higher recorded model-construction cost. Thus the contribution
is an explicit, interpretable description of the missing
sampling interaction, with measurable predictive and selection
consequences, rather than dominance over every numerical baseline.

\resultlead{Scope}
The expansion assumes the specified bilinear sampler and local
transformation geometry. Transfer to realized network margins
requires local regularity, and the unrestricted-additive
obstruction assumes independent transformation coordinates.
Exact surrogate optimization and conditional margin-regret bounds
do not establish finite-scale robustness certification for the
nonlinear network. The multiclass experiment uses an enriched
image cohort and a fixed competitor shortlist; its recorded
construction times are not a matched end-to-end benchmark.
Within these limits, the results connect a concrete numerical
sampling mechanism to margin fidelity, decision prediction, and
the choice of joint transformations that challenge a classifier.

\FloatBarrier


\begin{thebibliography}{99}
\setlength{\itemsep}{-1.4pt}
\setlength{\parsep}{0pt}
\footnotesize

\bibitem{simard1991}
P. Simard, B. Victorri, Y. LeCun, J. Denker,
Tangent Prop: A formalism for specifying selected invariances in an adaptive network,
in: NeurIPS 4 (1991) 895--903.

\bibitem{simard2000}
P.Y. Simard, Y.A. LeCun, J.S. Denker, B. Victorri,
Transformation invariance in pattern recognition: Tangent distance and propagation,
Int. J. Imaging Syst. Technol. 11 (2000) 181--197.

\bibitem{jaderberg2015}
M. Jaderberg, K. Simonyan, A. Zisserman, K. Kavukcuoglu,
Spatial Transformer Networks,
in: NeurIPS 28, 2015, pp. 2017--2025.


\bibitem{shapiro1990}
A. Shapiro,
On concepts of directional differentiability,
J. Optim. Theory Appl. 66 (1990) 477--487.
https://doi.org/10.1007/BF00940933.

\bibitem{fangsantos2019}
Z. Fang, A. Santos,
Inference on directionally differentiable functions,
Rev. Econ. Stud. 86 (1) (2019) 377--412.
https://doi.org/10.1093/restud/rdy049.


\bibitem{engstrom2019}
L. Engstrom, B. Tran, D. Tsipras, L. Schmidt, A. Madry,
Exploring the landscape of spatial robustness,
in: ICML, PMLR 97 (2019) 1802--1811.

\bibitem{polavarapu1996}
S. Polavarapu, M. Tanguay, R. M\'enard, A. Staniforth,
The tangent linear model for semi-Lagrangian schemes: Linearizing the process of interpolation,
Tellus A 48 (1996) 74--95.

\bibitem{tanguay1997}
M. Tanguay, S. Polavarapu, P. Gauthier,
Temporal accumulation of first-order linearization error for semi-Lagrangian passive advection,
Mon. Weather Rev. 125 (1997) 1296--1311.

\bibitem{scholtes2012}
S. Scholtes,
Introduction to Piecewise Differentiable Equations,
Springer, 2012.

\bibitem{griewank2013}
A. Griewank,
On stable piecewise linearization and generalized algorithmic differentiation,
Optim. Methods Softw. 28(6) (2013) 1139--1178.

\bibitem{balunovic2019}
M. Balunovi\'c, M. Baader, G. Singh, T. Gehr, M. Vechev,
Certifying geometric robustness of neural networks,
in: NeurIPS 32, 2019, pp. 15287--15297.


\bibitem{rabitzalis1999}
H. Rabitz, \"O.F. Ali\c{s},
General foundations of high-dimensional model representations,
J. Math. Chem. 25(2--3) (1999) 197--233.
https://doi.org/10.1023/A:1019188517934.


\bibitem{fischer2020}
M. Fischer, M. Baader, M. Vechev,
Certified defense to image transformations
via randomized smoothing,
in: NeurIPS 33, 2020.


\bibitem{buhmann1999}
M.D. Buhmann, A. Pinkus,
Identifying linear combinations of ridge functions,
Adv. Appl. Math. 22(1) (1999) 103--118.
https://doi.org/10.1006/aama.1998.0623.

\bibitem{pinkus2013}
A. Pinkus,
Smoothness and uniqueness in ridge function representation,
Indag. Math. (N.S.) 24(4) (2013) 725--738.
https://doi.org/10.1016/j.indag.2012.10.004.

\bibitem{jiang2019}
W. Jiang, W. Sun, A. Tagliasacchi, E. Trulls, K.M. Yi,
Linearized multi-sampling for differentiable image transformation,
in: ICCV, 2019, pp. 2988--2997.

\bibitem{batten2024}
B. Batten, Y. Zheng, A. De Palma, P. Kouvaros, A. Lomuscio,
Verification of geometric robustness of neural networks via piecewise linear approximation and Lipschitz optimisation,
in: ECAI, 2024, pp. 2362--2369.

\bibitem{krizhevsky2009}
A. Krizhevsky,
Learning multiple layers of features from tiny images,
Technical Report, University of Toronto, 2009.

\bibitem{wolberg1990}
G. Wolberg,
\textit{Digital Image Warping},
IEEE Computer Society Press, Los Alamitos, CA, 1990.

\bibitem{he2016}
K. He, X. Zhang, S. Ren, J. Sun,
Deep residual learning for image recognition,
in: CVPR, 2016, pp. 770--778.

\bibitem{zagoruyko2016}
S. Zagoruyko, N. Komodakis,
Wide Residual Networks,
in: BMVC, 2016, pp. 87.1--87.12.

\bibitem{netzer2011}
Y. Netzer, T. Wang, A. Coates, A. Bissacco,
B. Wu, A.Y. Ng,
Reading digits in natural images with unsupervised feature learning,
in: NIPS Workshop, 2011.

\bibitem{recht2019imagenetv2}
B. Recht, R. Roelofs, L. Schmidt, V. Shankar,
Do ImageNet classifiers generalize to ImageNet?,
in: ICML, PMLR 97 (2019) 5389--5400.

\bibitem{liu2022convnext}
Z. Liu, H. Mao, C.-Y. Wu, C. Feichtenhofer,
T. Darrell, S. Xie,
A ConvNet for the 2020s,
in: CVPR, 2022, pp. 11976--11986.

\bibitem{riguet1951}
J. Riguet,
Les relations de Ferrers,
C. R. Acad. Sci. Paris 232 (1951) 1729--1730.

\bibitem{wang2023georobust}
F. Wang, P. Xu, W. Ruan, X. Huang,
Towards verifying the geometric robustness of large-scale neural networks,
in: AAAI 37(12), 2023, pp. 15197--15205.
https://doi.org/10.1609/aaai.v37i12.26773.

\bibitem{cohen2026superdeepg}
N. Cohen, M. Ducoffe, T. Fayard, C. Gabreau, C. Pagetti, X. Pucel,
Certified geometric robustness---Super-DeepG,
in: Proc. 17th ACM/IEEE International Conference on Cyber-Physical Systems (ICCPS), 2026, pp. 242--252.
https://doi.org/10.1109/ICCPS68698.2026.00031.

\bibitem{lixiu2010}
J. Li, D. Xiu,
Evaluation of failure probability via surrogate models,
J. Comput. Phys. 229 (2010) 8966--8980.


\bibitem{hoeffding1948}
W. Hoeffding,
A class of statistics with asymptotically normal distribution,
Ann. Math. Statist. 19(3) (1948) 293--325.
https://doi.org/10.1214/aoms/1177730196.

\bibitem{aliev2024}
R.A. Aliev, F.M. Isgandarli,
On the representability of a continuous multivariate function by
sums of ridge functions,
J. Approx. Theory 304 (2024) 106105.
https://doi.org/10.1016/j.jat.2024.106105.

\bibitem{freixas2017}
J. Freixas, M. Freixas, S. Kurz,
On the characterization of weighted simple games,
Theory Decis. 83 (2017) 469--498.
https://doi.org/10.1007/s11238-017-9606-z.

\end{thebibliography}
\end{document}